\documentclass[conference,compsoc,anonymous]{IEEEtran}
\ifCLASSOPTIONcompsoc
  \usepackage[nocompress]{cite}
\else
  \usepackage{cite}
\fi
\ifCLASSINFOpdf
  \usepackage[pdftex]{graphicx}
  \usepackage{multirow}
  \usepackage{hyperref}
  \graphicspath{{../pdf/}{../jpeg/}}
  \DeclareGraphicsExtensions{.pdf,.jpeg,.png}

\usepackage{amsmath} 
\usepackage{algorithm} 
\usepackage{algpseudocode} 

\usepackage{array}
\usepackage{nicefrac}
\usepackage{siunitx}
\usepackage{array,framed}
\usepackage{booktabs}
\usepackage{
  color,
  float,
  epsfig,
  wrapfig,
  graphics,
  graphicx,
  subcaption
}
\AtEndPreamble{
\usepackage{hyperref}
\hypersetup{
    colorlinks = true,
    linkcolor = red,
    anchorcolor = red,
    citecolor = red,
    filecolor = red,
    urlcolor = red
}
}
\usepackage{textcomp,amssymb}
\usepackage{setspace}
\usepackage{latexsym,fancyhdr,url}
\usepackage{enumerate}
\usepackage{graphics}
\usepackage[framemethod=TikZ]{mdframed}
\usepackage{xparse} 
\usepackage{xspace}
\usepackage{multirow}
\usepackage{csvsimple}
\usepackage{balance}
\newcommand{\myframed}[2]{%
  \begin{mdframed}[linecolor=#1!90,backgroundcolor=#1!4,roundcorner=2pt,linewidth=1.5pt]
  #2
  \end{mdframed}
}

\usepackage{
  tikz,
  pgfplots,
  pgfplotstable
}
\usepackage{hyperref}
\definecolor{deepseek-r}{RGB}{91, 181, 172}    
\definecolor{deepseek-nr}{RGB}{139, 218, 210}   
\definecolor{qwen-r}{RGB}{216, 179, 101}        
\definecolor{qwen-nr}{RGB}{235, 203, 136}       
\definecolor{llama-r}{RGB}{222, 82, 108}        
\definecolor{llama-nr}{RGB}{243, 126, 148}      

\usepackage{tabularx}
\usepackage{booktabs}
\usepackage{caption}
\usepackage{subcaption}
\usepackage{newtxmath}
\usepackage{threeparttable}

\usepackage{bm} 
\usepackage{xcolor}
\usepackage{colortbl}
\usepackage{lipsum} 
\usepackage{multicol}
\usepackage[utf8]{inputenc}
\usepackage{pifont}
\usepackage{tikz}
\usepackage{placeins}
\usepackage{afterpage}
\usepackage{makecell}

\definecolor{lightpink}{rgb}{1.0, 0.95, 0.95}
\definecolor{lightyellow}{rgb}{1.0, 1.0, 0.88}
\definecolor{lightgreen}{rgb}{0.88, 1.0, 0.88}
\definecolor{lightgray}{rgb}{0.9, 0.9, 0.9}
\definecolor{mygreen}{RGB}{34,139,34}

\usepackage[most]{tcolorbox}
\usepackage{tcolorbox}
\newcounter{findingcounter}
\newtcolorbox[auto counter]{myboxRQ}[2][]{%
    enhanced jigsaw,
    colframe=black,
    colback=white, 
    boxrule=0.6pt, 
    arc=0pt, 
    unbreakable,
    fonttitle=\bfseries,
    top=1.2pt,
    bottom=1.2pt,
    #1
}

\newcommand{\findbox}[1]{%
  \stepcounter{findingcounter} 
  \myframed{qwen-nr}{\textbf{Finding \#\thefindingcounter: }#1} 
}
\newcommand{\MethodName}{RELIANCE \xspace}
\newcommand{\MethodNameNoSpace}{RELIANCE\xspace}

\begin{document}
%
\title{Trustworthy Reasoning: Evaluating and Enhancing Factual Accuracy in LLM Intermediate Thought Processes}



%
\author{\IEEEauthorblockN{Rui Jiao\IEEEauthorrefmark{1},
Yue Zhang\IEEEauthorrefmark{2},
Jinku Li\IEEEauthorrefmark{1}}
\IEEEauthorblockA{\IEEEauthorrefmark{1}School of Cyber Engineering, Xidian University}
\IEEEauthorblockA{\IEEEauthorrefmark{2}School of Computer Science and Technology, Shandong University}
}


\newenvironment{packed_enum}{
\begin{enumerate}
  \setlength{\itemsep}{0pt}
  \setlength{\parskip}{0pt}
  \setlength{\parsep}{0pt}
}{\end{enumerate}}

\newenvironment{packed_item}{
\begin{itemize}
  \setlength{\itemsep}{1pt}
  \setlength{\parskip}{0pt}
  \setlength{\parsep}{0pt}
}{\end{itemize}}
\newenvironment{packeditemize}{
\begin{list}{$\bullet$}{
\setlength{\itemsep}{1.5pt}
\setlength{\labelwidth}{8pt}
\setlength{\leftmargin}{10pt}
\setlength{\labelsep}{3pt}
\setlength{\listparindent}{\parindent}
\setlength{\parsep}{1.5pt}
\setlength{\parskip}{1.5pt}
\setlength{\topsep}{1.5pt}}}{\end{list}}

\def \sysname{\textsf{Santoryu}\xspace}

\maketitle

\begin{abstract}

We present a novel framework addressing a critical vulnerability in Large Language Models (LLMs): the prevalence of factual inaccuracies within intermediate reasoning steps despite correct final answers. This phenomenon poses substantial risks in high-stakes domains including healthcare, legal analysis, and scientific research, where erroneous yet confidently presented reasoning can mislead users into dangerous decisions. Our framework integrates three core components: (1) a specialized fact-checking classifier trained on counterfactually augmented data to detect subtle factual inconsistencies within reasoning chains; (2) an enhanced Group Relative Policy Optimization (GRPO) reinforcement learning approach that balances factuality, coherence, and structural correctness through multi-dimensional rewards; and (3) a mechanistic interpretability method examining how factuality improvements manifest in model activations during reasoning processes. Extensive evaluation across multi state-of-the-art models reveals concerning patterns: even leading models like Claude-3.7 and GPT-o1 demonstrate reasoning factual accuracy of only 81.93\% and 82.57\% respectively. Our approach significantly enhances factual robustness (up to 49.90\% improvement) while maintaining or improving performance on challenging benchmarks including Math-500, AIME-2024, and GPQA. Furthermore, our neural activation-level analysis provides actionable insights into how factual enhancements reshape reasoning trajectories within model architectures, establishing foundations for future training methodologies that explicitly target factual robustness through activation-guided optimization.
\looseness=-1
\end{abstract}


\section{Introduction}

Large language models (LLMs) have demonstrated remarkable capabilities in reasoning and problem-solving across diverse domains~\cite{DBLP:conf/nips/BrownMRSKDNSSAA20,DBLP:journals/corr/abs-2212-08073,DBLP:journals/corr/abs-2302-13971,DBLP:journals/corr/abs-2312-11805,vicuna2023,qwen}.
Among them, reasoning LLMs represent a distinguished subclass of LLMs. Their defining characteristic is the generation of intermediate reasoning steps followed by a final answer when responding to user queries. This reasoning content, also produced by the model, is typically enclosed within special tags, and the final answer is appended immediately after the reasoning (e.g., \textit{``$<$think$>$reasoning process$<$/think$>$final answer''}).
Many reasoning LLMs yields even more impressive performance~\cite{DBLP:journals/corr/abs-2501-12948,qwq32b,OpenAIo1,Claude-3.7}.
However, these reasoning LLMs in high-stakes scenarios, e.g., medical diagnosis~\cite{DBLP:journals/corr/abs-2409-00097,DBLP:journals/corr/abs-2402-01730}, legal reasoning~\cite{DBLP:journals/corr/abs-2502-17638,DBLP:journals/corr/abs-2502-05675}, and financial decision-making,  demand~\cite{DBLP:journals/corr/abs-2412-18174,DBLP:conf/icaif/SinghKS24} not just convincing answers but factually accurate reasoning processes. While significant progress has been made in improving the final outputs of LLMs, the intermediate reasoning steps—the ``thinking process'' that leads to conclusions—often contain critical factual errors that undermine the reliability and trustworthiness of these systems~\cite{DBLP:journals/corr/abs-2408-05093}.\looseness=-1

These factual errors partly stem from the Reinforcement Learning (RL) training process, which encourages the model to explore reasoning steps through rewards and generate final answers accordingly. While effective at making responses more human-like and engaging, RL can inadvertently encourage models to ``pretend to know'' when they do not, fabricating plausible-sounding but incorrect explanations to satisfy perceived expectations. Additionally, specialized training in mathematical or logical reasoning, intended to improve step-by-step problem solving, may reinforce this behavior. Once a mistake is introduced in the reasoning steps chain, models often generate self-consistent but fundamentally flawed justifications rather than acknowledging uncertainty or correcting errors.\looseness=-1

This issue is critical: factual errors introduced early in the visible reasoning process can propagate and amplify, ultimately leading to incorrect conclusions. More concerningly, when these intermediate thinking steps are explicitly presented to users in a coherent but factually flawed structure, the thinking model's erroneous reasoning appears more convincing and harder for users to detect, despite containing critical factual errors.
For instance, when asked a reasoning LLM \textit{``Can morphine be used to treat vomiting in a one-year-old child ?''}, an LLM might generate a reasoning process that appears logical on the surface but is clinically dangerous. It might begin by noting that \textit{``morphine is used to treat pain -$>$ morphine could be used to treat vomiting -$>$ then proceed to calculate a dose based on body weight: concluding that 0.1 mg per kilogram is appropriate -$>$ leading to a dose of 0.22 mg for a one-year-old. -$>$ If symptoms persist, it may even suggest increasing the dose to 0.4 mg.'' (``-$>$''  represents a step in a reasoning chain, indicating a cause-and-effect)} Although this reasoning may seem methodical and authoritative, it contains serious and potentially life-threatening errors: \textit{morphine is not indicated for pediatric vomiting, it can cause respiratory depression in infants, and the suggested dosing strategy is entirely inappropriate}.

Beyond the aforementioned medical advice issue, factual inconsistencies in legal reasoning and contract analysis may expose individuals or organizations to legal and financial liabilities. Misinterpretation of statutes or precedents by LLMs can result in misguided legal actions. Similarly, in scientific inquiry and hypothesis generation, factual inaccuracies from LLMs can compromise the epistemic integrity of scientific discourse and contribute to large-scale misinformation. User trust in LLMs hinges on the factual reliability of their observable thinking processes. Encountering a single significant factual error within the displayed reasoning chain can substantially erode user confidence in the entire system, even in domains where previous outputs were accurate. This phenomenon of ``trust collapse'' highlights the urgency of enhancing factual accuracy throughout the visible thinking process of LLMs. To make matters worse,  malicious users may intentionally design prompts to inject false information, crafting misleading premises or contextual cues to induce the model to generate factually incorrect yet internally consistent reasoning chains.\looseness=-1

Current fact-checking methods exhibit several critical limitations. First, most approaches focus on final-answer level verification rather than evaluating the full thinking process, overlooking the compounding effect of intermediate factual errors that users directly observe~\cite{DBLP:conf/acl/0002ZKY0GDB24,DBLP:conf/acl/LinHE22,DBLP:conf/nips/HuangBZZZSLLZLF23}.
Second, there is a lack of effective mechanisms for factual correction that preserve the coherence of the visible thinking process~\cite{DBLP:conf/nips/LeePXPFSC22,DBLP:conf/naacl/MoiseevDAJ22,DBLP:conf/aaai/SunSGRRR23}. 
Third, existing methods offer limited interpretability, failing to expose how factual errors emerge and propagate within the thinking model's observable reasoning steps that significantly impact user trust and decision-making.
These limitations motivate the design of a unified framework that detects and corrects factual errors throughout the observable thinking process and employs mechanistic interpretability techniques to analyze how thinking models generate these reasoning steps, thereby enhancing the factual robustness of the user-facing reasoning that ultimately determines trust and utility in practical applications.\looseness=-1

Therefore, in this paper, we aim to enhance the factual robustness of the observable reasoning steps in LLMs and builds user trust through consistently accurate reasoning chains. ~\autoref{tab:hallucination_comparison} illustrates the distinction between our research problem and the general hallucination problem.
Particularly, we focus on three research questions: 
\begin{itemize}
    \item [\textbf{RQ1}] What is the current landscape of factuality in reasoning steps of state-of-the-art LLMs?
    \item [\textbf{RQ2}] How can we enhance factual grounding in reasoning steps while promoting appropriate acknowledgment of knowledge limitations in LLMs?
    \item [\textbf{RQ3}] How does factuality enhancement training affect the internal neural activations of LLMs reasoning steps, and are there methods to support these insights to guide more factually robust model training in the future?
\end{itemize}

To address these questions, we introduce \MethodName, a comprehensive framework that integrates:
(1) For RQ1, we construct factual and counterfactual reasoning dataset using Natural Language Processing (NLP) techniques and train a specialized fact-checking classifier via supervised fine-tuning (SFT). We then evaluate the factual consistency of reasoning steps across multiple models, including both black-box and white-box LLMs. 
(2) For RQ2, we propose a reinforcement learning-based factuality enhancement method employing Group Relative Policy Optimization (GRPO)~\cite{DBLP:journals/corr/abs-2402-03300}, and carefully designed multi-dimensional reward signals to improve the factual consistency of reasoning in LLMs. We adopt GRPO as our reinforcement learning framework because it is specifically designed to optimize structured reasoning in language models. 
(3) For RQ3, we introduce a mechanistic interpretability method~\cite{DBLP:journals/corr/abs-2404-14082} to analyze model latent activations during reasoning.

Experimental results demonstrate that our method identifies factual errors in the reasoning steps of both four white-box and six black-box LLMs (RQ1), for example, even state-of-the-art models such as \textit{Claude-3.7}~\cite{Claude-3.7} and \textit{GPT-o1}~\cite{OpenAIo1} achieve only 81.93\% and 82.57\% factual accuracy in their reasoning process steps, respectively. 
Our proposed factuality enhancement algorithm significantly improves the factual consistency of reasoning steps (RQ2), achieving an accuracy gain of up to 49.90\%. 
Finally, the mechanistic interpretability technique reveals how factuality improvements correspond to changes in latent neural activations, providing insights for future training algorithm design, designing training objectives to strengthen inter-step dependencies and emphasize the activation of ``aha moments'' can further enhance model reasoning steps factuality (RQ3).

In summary, our main contributions include:
\begin{itemize}
    \item We conduct a comprehensive evaluation of reasoning LLMs, revealing significant factual errors in reasoning steps across state-of-the-art models.
    
    \item We develop a reinforcement learning-based factuality enhancement framework that improves reasoning factuality by up to 49.90\% while preserving final output performance on standard benchmarks.
    
    \item We employ mechanistic interpretability techniques to analyze how factuality improvements manifest in neural activations, providing insights to guide future factually robust model training.
\end{itemize}


\section{Background and Related Work}


\begin{table}[t]
\centering
\small 
\caption{Comparison Between Hallucinations and LLM Reasoning Steps Factuality}
\label{tab:hallucination_comparison}
\resizebox{\columnwidth}{!}{%
\begin{tabular}{@{}p{8.5cm}p{2.0cm}p{1.5cm}@{}}
\toprule
\textbf{Characteristic} & \textbf{Hallucinations} & \textbf{Reasoning Steps Factuality} \\
\midrule
Avoidance of fabricated details or entities & \ding{55} & \ding{51} \\
\addlinespace[0.2em]
Reasoning steps both internally and externally correct & \ding{55} & \ding{51} \\
\addlinespace[0.2em]
Appropriate expression of uncertainty when knowledge is limited & \ding{55} & \ding{51} \\
\addlinespace[0.2em]
Prevention of error propagation through reasoning chain & \ding{55} & \ding{51} \\
\addlinespace[0.2em]
Alignment with domain-specific expertise & \ding{55} & \ding{51} \\
\addlinespace[0.2em]
Avoidance of plausible-sounding but unfounded reasoning & \ding{55} & \ding{51} \\
\addlinespace[0.2em]
Accuracy in numerical calculations and statistics & \ding{55} & \ding{51} \\
\addlinespace[0.2em]
Recognition of knowledge boundaries & \ding{55} & \ding{51} \\
\addlinespace[0.2em]
Capability for self-correction when errors are detected & \ding{55} & \ding{51} \\
\bottomrule
\end{tabular}%
}
\begin{tablenotes}
\small
\item \ding{51} = Present/Yes; \ding{55} = Absent/No
\end{tablenotes}
\end{table}

\noindent\textbf{Factuality Issues in LLMs.} 
The factuality issue in LLMs is defined as the probability of these models producing content inconsistent with established facts~\cite{DBLP:conf/fat/BenderGMS21,DBLP:journals/corr/abs-2303-12712,DBLP:journals/corr/abs-2303-08774}. This problem is particularly concerning given the widespread deployment of LLMs in applications such as search engines~\cite{MicrosoftBing2023}, chatbots~\cite{GoogleBard2023,characterAI,POE}, and content generators~\cite{DBLP:journals/corr/abs-2306-16092}. While hallucination and factuality issues are related, they address distinct aspects: hallucination primarily refers to the model's tendency to produce baseless or unwarranted content, whereas factuality concerns its ability to produce content that is irrelevant to or contradicts established facts~\cite{DBLP:journals/csur/JiLFYSXIBMF23,DBLP:journals/corr/abs-2303-08774}.
Recent studies have categorized factual errors in LLMs into several types: (1) domain knowledge deficit, where models lack expertise in specific domains~\cite{DBLP:conf/nips/LuMX0CZTCK22}; (2) outdated information, where models are unaware of recent developments~\cite{DBLP:journals/corr/abs-2112-09332,DBLP:conf/acl/QinCJYLZLHDWXQL23}; (3) immemorization, where models fail to retain knowledge from their training corpus; (4) reasoning failures, where models possess knowledge but fail to reason with it effectively~\cite{DBLP:conf/iclr/BerglundTKBSKE24,DBLP:conf/iclr/KothaSR24}; and (5) exposure bias, where models reflect training data biases rather than objective factuality~\cite{DBLP:journals/corr/abs-2309-00770,DBLP:conf/acl/HossainD023}.

\vspace{1mm}
\noindent\textbf{Strategies for Enhancing Factuality.} 
To better understand the limitations of current evaluation methodologies for factuality in LLMs, it is crucial to recognize that these approaches often focus on the final output's accuracy, neglecting the underlying reasoning steps that lead to these outputs. Most evaluation methods, such as MMLU~\cite{DBLP:conf/acl/0002ZKY0GDB24}, TruthfulQA~\cite{DBLP:conf/acl/LinHE22}, and C-Eval~\cite{DBLP:conf/nips/HuangBZZZSLLZLF23}, assess the factual correctness of a model's final answer but do not scrutinize the intermediate reasoning processes that lead to these answers. While these benchmarks offer valuable insights, they fail to address how factually accurate the reasoning steps are, creating a significant gap in our understanding of the consistency and reliability of the entire reasoning process.

In light of this gap, research has been focusing on ways to enhance the factuality of LLMs, with efforts primarily falling into two broad categories: improving the standalone capabilities of LLMs and augmenting them with external knowledge sources. For standalone LLMs, various strategies such as continual pretraining with topic prefixes~\cite{DBLP:conf/nips/LeePXPFSC22}, supervised fine-tuning~\cite{DBLP:conf/naacl/MoiseevDAJ22,DBLP:conf/aaai/SunSGRRR23}, and model editing~\cite{DBLP:conf/acl/DaiDHSCW22,DBLP:conf/nips/MengBAB22} have been explored. Moreover, multi-agent approaches~\cite{DBLP:conf/icml/Du00TM24,DBLP:conf/emnlp/CohenHGG23}, where multiple language model instances debate to reach consensus, have shown promise in improving factual accuracy. Additionally, new prompting techniques~\cite{DBLP:conf/iclr/0002IWXJ000023,DBLP:conf/eacl/WellerMWLKD24} and decoding strategies~\cite{DBLP:conf/nips/LeePXPFSC22,DBLP:conf/iclr/ChuangXLKGH24} aim to refine the model’s reasoning process without altering its parameters. 
On the other hand, retrieval-augmented LLMs (RAG) introduce external knowledge as a tool to enhance factuality. Techniques include retrieving information from external sources~\cite{DBLP:conf/icml/BorgeaudMHCRM0L22,DBLP:journals/corr/abs-2203-05115}, interactive retrieval methods~\cite{DBLP:journals/corr/abs-2301-00303,DBLP:conf/acl/TrivediBKS23}, and retrieval adaptation strategies~\cite{DBLP:journals/jmlr/IzacardLLHPSDJRG23,DBLP:conf/naacl/ShiMYS0LZY24,DBLP:journals/corr/abs-2305-15225}. Moreover, recent work has explored retrieving from external memory and structured knowledge sources~\cite{DBLP:conf/nips/LiGK22,DBLP:conf/emnlp/WanYZSSCJL22,DBLP:journals/corr/abs-2305-13669} to further improve factual accuracy.

While these methods have shown promising results in enhancing the factuality of final outputs, they primarily address the output's accuracy rather than the factual integrity of the reasoning process. Furthermore, many of these approaches rely on external knowledge, which may not always be available or reliable in all scenarios, limiting their general applicability. This underscores the need for a shift in focus toward improving the model’s inherent factual reasoning abilities, especially when external knowledge sources are scarce or unreliable.

\section{Design of \MethodName}

\begin{figure*}[t!]
    \centering
    \includegraphics[trim=0 0 0 0, clip, width=\textwidth]{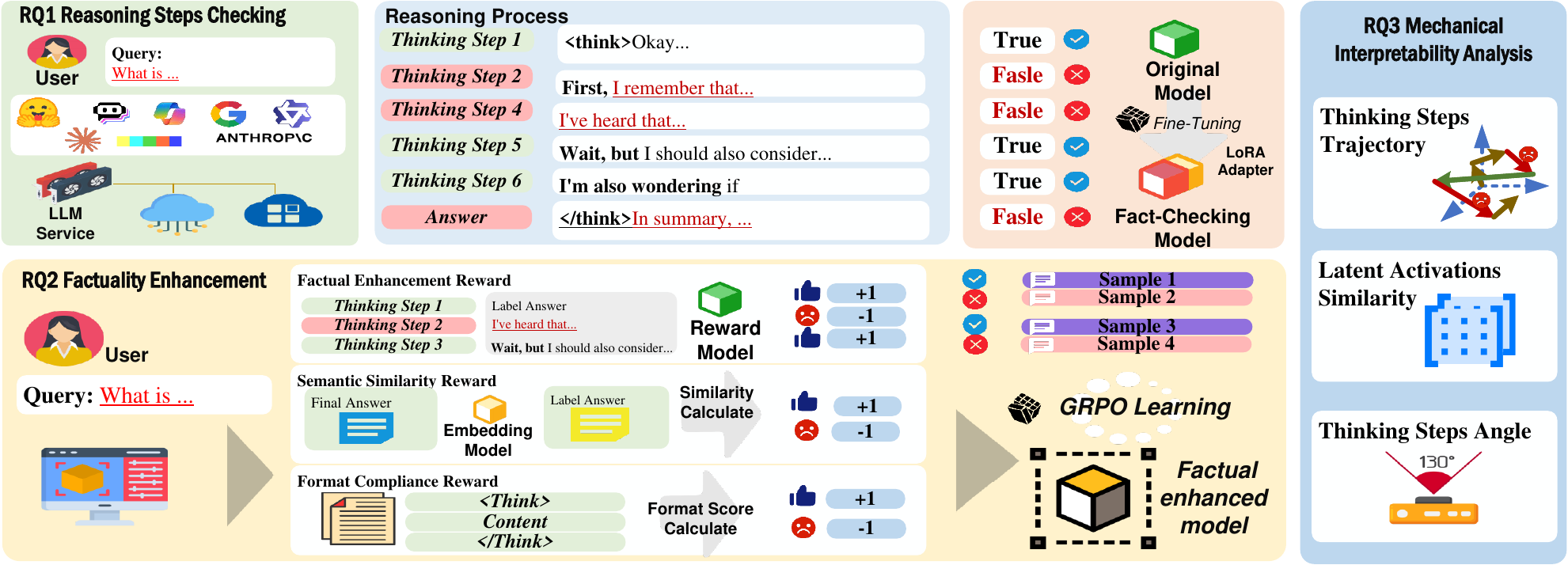}
    \caption{The workflow of \MethodName.}
    \label{fig:Arc_fig}
\end{figure*}

We present our comprehensive framework for fact-checking in LLM reasoning steps and enhancing factuality throughout the reasoning process. We named our framework \MethodName (\textbf{R}easoning \textbf{E}valuation with \textbf{L}ogical \textbf{I}ntegrity and \textbf{A}ccuracy for \textbf{C}onfidence \textbf{E}nhancement).
The framework consists of three integrated components: 
a specialized fact-checking classifier trained on counterfactually-augmented reasoning data (addressing \hyperlink{rq1}{RQ1}); a reinforcement learning-based enhancement mechanism using Group Relative Policy Optimization (GRPO) with multi-faceted rewards (addressing \hyperlink{rq2}{RQ2}); and a mechanical interpretability method that analyzes model neural activations during reasoning (addressing \hyperlink{rq3}{RQ3}).
We first introduce our fact-checking methodology in §~\ref{subsec:fact-checking}, then present our reasoning steps factuality enhancement method in §~\ref{subsec:enhancement}, and finally describe our mechanical interpretability analysis in  §~\ref{subsec:interpretability}.

\subsection{Fact-Checking for Reasoning Steps (RQ1)}
\label{subsec:fact-checking}

As previously highlighted, our goal is to understand the current landscape of factuality in the reasoning steps of state-of-the-art LLMs. To achieve this, we propose evaluating factual accuracy in LLM reasoning by leveraging a specialized fact-checking pipeline. Our methodology addresses the challenge of assessing factual accuracy within complex reasoning process steps by developing a classifier trained on counterfactually augmented data. We establish a comprehensive pipeline that includes data collection from reliable sources, systematic entity manipulation for counterfactual generation, and supervised fine-tuning to detect factual inconsistencies specifically within chain-of-thought reasoning steps.

\vspace{1mm}
\noindent{\textbf{Phase I, Data Collection and Preprocessing:}}
We collected over 20,000 Wikipedia entries as our factual reference base due to their reliability, domain coverage, and structured information. From this corpus, we extracted approximately 6,000 samples containing diverse named entities (e.g., person names, locations, organizations) to train an LLM-based classifier capable of identifying factual inconsistencies across various knowledge domains. Using the \textit{DeepSeek-R1-671B} API
~\cite{DBLP:journals/corr/abs-2501-12948}, we transformed these samples into question-answer (Q\&A) pairs with explicit reasoning steps, resulting in 6,000 CoT~\cite{DBLP:conf/nips/Wei0SBIXCLZ22} reasoning Q\&A pairs that preserve factual integrity while providing explicit reasoning steps for evaluation.

To rigorously evaluate a model’s ability to detect factual errors in multi-step reasoning, we require a dataset that contains both factually correct and subtly corrupted reasoning chains. However, naturally occurring factually incorrect reasoning data is scarce and often lacks ground truth labels. To address this, we introduce a controlled method for injecting factual errors into existing reasoning samples while preserving their grammatical and contextual plausibility.
To facilitate controlled entity manipulation, we first applied the \textit{``flair/ner-english-ontonotes-large''}~\cite{akbik2019flair} model for Named Entity Recognition (NER), identifying over 46,000 unique entities spanning multiple categories. 
Detailed category information is provided in Appendix \ref{appendix-NERCategory}.
We then implemented systematic entity substitution on the original CoT reasoning samples, randomly replacing entities with different entities of the same type to introduce factual errors while maintaining grammatical coherence. The original unmodified samples were labeled as positive instances (factually correct), while samples with substituted entities were labeled as negative instances (containing factual errors). This procedure yielded 38,539 training samples, each consisting of a question, reasoning steps, and binary factuality label. We allocated 1,000 samples each for test and validation sets, ensuring no overlap with training data. \autoref{tab:entity_replacement} shows an example instance from our dataset.

\begin{table}[t]
\centering
\scriptsize 
\caption{Example of our named entity replacement method used in the data augmentation pipeline.}
\label{tab:entity_replacement}
\begin{tabular}{p{3.0cm}p{2.5cm}p{2cm}}
\toprule
\textbf{Original Entity} & \textbf{Replacement} & \textbf{Entity Type} \\
\midrule
Adelaide Anne Procter & Gene Wilder & PERSON \\
\midrule
\multicolumn{3}{p{7.5cm}}{\textbf{Original Text}: \textit{Adelaide Anne Procter} (30 October 1825 -- 2 February 1864) was a significant English poet and philanthropist, known for her active role in social causes...} \\
\midrule
\multicolumn{3}{p{7.5cm}}{\textbf{Transformed Text}: \textit{Gene Wilder} (30 October 1825 -- 2 February 1864) was a significant English poet and philanthropist, known for her active role in social causes...} \\
\bottomrule
\end{tabular}
\end{table}

\vspace{1mm}
\noindent{\textbf{Phase II, Training:}}
Building on the previously constructed dataset of factuality reasoning steps, we train LLMs to enable fine-grained detection of factual correctness within reasoning steps. We formalize our Supervised Fine-Tuning (SFT)~\cite{DBLP:conf/nips/BrownMRSKDNSSAA20} method for training the fact-checking classifier. 
{We adopt SFT to train the fact-checking classifier, as LLMs are initially pretrained for continuation tasks. Similar to how instruction-tuned models are aligned with Q\&A tasks through SFT, we leverage SFT to adapt the model to a binary factuality classification task. SFT has been widely used for domain-specific or task-specific adaptation~\cite{DBLP:journals/corr/abs-2308-10792}.}
Additionally, the structured nature of our labeled dataset makes SFT particularly well-suited for developing a reliable fact-checking mechanism that can generalize across diverse reasoning contexts. The core objective is to optimize the model's ability to discriminate between factually correct and erroneous reasoning steps.  \looseness=-1

As shown in Part 1 of~\autoref{fig:Arc_fig}, \textit{Step 2, Step 4, and the final answer} contain factual errors. 
We train the model's LoRA adapter~\cite{DBLP:conf/iclr/HuSWALWWC22} to enable the LLM to identify such fine-grained errors. 
{We adopt LoRA primarily due to its widespread use in LLM fine-tuning and its superior performance over full-parameter training on certain specialized tasks. LoRA also mitigates catastrophic forgetting~\cite{DBLP:journals/corr/abs-2405-09673}, a common issue in full-parameter training.}
Specifically, given our dataset $\mathcal{D} = \{(x_i, y_i)\}_{i=1}^n$ where each sample consists of an input $x_i$ (comprising the system prompt, question, and reasoning chain steps) and a corresponding factuality label $y_i$ (either ``$<$fact$>$True$<$/fact$>$'' or ``$<$fact$>$False$<$/fact$>$''), we implement a standard autoregressive language modeling objective. The loss function is defined as:

\begin{equation}
\mathcal{L}_{\text{SFT}}(\theta) = -\frac{1}{n} \sum_{i=1}^n \sum_{j=1}^{|y_i|} \log P_\theta(y_{i,j} | x_i, y_{i,<j})
\end{equation}

\noindent
where the $\theta$ denotes the model parameters
$P_\theta(y_{i,j} | x_i, y_{i,<j})$ represents the probability the model assigns to the $j$-th token of output $y_i$ conditioned on input $x_i$ and preceding output tokens $y_{i,<j}$,
the $|y_i|$ indicates the length of the target sequence.
During optimization, we minimize this loss function using the AdamW optimizer~\cite{DBLP:conf/iclr/LoshchilovH19} coupled with a cosine learning rate scheduler incorporating warmup. 
The input representation for each instance is constructed as:

\begin{equation}
x_i = \text{[system-prompt]} \oplus \text{[cot-question]} \oplus \text{[cot-answer]}
\end{equation}

\noindent
with target outputs formulated as:

\begin{equation}
y_i \in \{\text{$<$fact$>$True$<$/fact$>$}, \text{$<$fact$>$False$<$/fact$>$}\}
\end{equation}

\noindent
This formulation enables the LLMs to learn the mapping between reasoning steps and factual correctness, ultimately developing the capacity to distinguish between valid reasoning steps and those containing factual errors.

\subsection{\textbf{Factuality Enhancement (RQ2)}}
\label{subsec:enhancement}

To bridge the gap between factual enhancement and high-quality reasoning in language models, we now extend our work beyond static fact-checking to actively enhancing reasoning through learning. 
While the previous section introduced a robust method for evaluating factual accuracy in each reasoning step, it does not directly inform the generation of such steps during inference. 
To address this limitation, we introduce a reinforcement learning method that leverages our fact-checking model as a key component of its reward function. Specifically, we adopt the GRPO (Group Relative Policy Optimization)~\cite{DBLP:journals/corr/abs-2402-03300} algorithm to guide the learning process, using a multi-faceted reward design that encourages factual consistency, semantic alignment, structural correctness, and appropriate verbosity.

We choose GRPO as our reinforcement learning backbone due to its tailored design for optimizing structured reasoning outputs in LLMs, particularly in reasoning-intensive tasks. Unlike traditional Proximal Policy Optimization (PPO)~\cite{DBLP:journals/corr/SchulmanWDRK17}, which evaluates outputs in isolation, GRPO introduces a group-based advantage computation that captures relative quality among a set of generated responses. This distinction is crucial in our setting, where the nuances of factual reasoning are best judged comparatively (e.g., which reasoning steps are more accurate or coherent under the same query context). Moreover, GRPO's incorporation of reference policy regularization mitigates reward hacking and helps maintain fluency and coherence by anchoring learning to a stable policy. These features make GRPO well-suited to integrate our multi-faceted reward components, enabling more effective and stable training of factually grounded reasoning policies.

\vspace{1mm}
\noindent{\textbf{Problem Formulation:}} We formalize our factuality enhancement task as a structured generation problem. Given a user query $q$, our goal is to generate a high-quality response $Re(q)$ consisting of both reasoning steps and a final answer:

\begin{equation}
Re(q) = \langle T(q), S(q) \rangle
\end{equation}

\noindent where the $T(q) = \{t_1, t_2, ..., t_n\}$ represents a sequence of reasoning steps, $S(q)$ denotes the final answer. Each reasoning step $t_i \in T(q)$ should maintain factual correctness while contributing to a coherent logical flow toward the final answer. 
For example, given a query \textit{``Who won the Nobel Prize in Physics the year after Albert Einstein?''}, a response might include, $t_1$: \textit{``Albert Einstein won the Nobel Prize in Physics in 1921.''}, $t_2$: \textit{``The Nobel Prize in Physics for 1922 was awarded to Niels Bohr.''}, $S$: \textit{``Niels Bohr won the Nobel Prize in Physics the year after Albert Einstein.''}
More detailed information, please refer to Appendix\ref{appendix-segmentation}.
Our objective is to train a policy $\pi_\theta$ that maximizes the probability of generating factually correct reasoning steps and answers, where $\theta$ represents the model parameters. Specifically, it can be defined as the following equation:

\begin{equation}
\pi_\theta^* = \arg\max_\theta \mathbb{E}_{q \sim \mathcal{D}} \left[ \sum_{T \in \mathcal{T}_q, S \in \mathcal{S}_q} \pi_\theta(T, S|q) \cdot R(T, S|q) \right]
\end{equation}

\noindent where $R(T, S|q)$ is our multi-faceted reward function that evaluates the quality of the generated reasoning steps and final answer.

\vspace{2mm}
\noindent{\textbf{Multi-faceted Reward Design:}}
While GRPO provides a strong foundation for optimizing reasoning answer quality in language models, directly applying it without modification falls short in our setting, where factuality is a central objective. Standard GRPO implementations typically rely on scalar rewards derived from generic quality signals such as human preferences or heuristic-based scores, which do not explicitly target factual correctness or structural fidelity. 
As a result, models trained under such rewards may generate fluent yet factually flawed reasoning steps. To address this limitation, we propose a task-specific modification of the GRPO reward function that incorporates multiple fine-grained criteria (i.e., factual correctness, semantic alignment, formatting consistency, and length appropriateness). 
By embedding our fact-checking model directly into the reward and explicitly encoding structural and semantic constraints, our adapted GRPO framework guides the model toward generating not only high-quality but also factually correct reasoning chains. 
The GRPO objective function is defined as the standard objective:



\begin{multline}
    \mathcal{J}_{\text{GRPO}} = 
    \mathbb{E}_{q \sim P(Q), \{o_i\}_{i=1}^G \sim \pi_{\theta_{\text{old}}}(O|q)} \bigg[ 
    \frac{1}{G} \sum_{i=1}^G \mathcal{M}_i \\
    - \beta \mathcal{D}_{\text{KL}} \big( \pi_\theta \| \pi_{\text{ref}} \big) \bigg]
    \label{eq:grpo}
\end{multline}

\begin{align}
    \mathcal{M}_i = \min \bigg( 
    \frac{\pi_\theta(o_i|q)}{\pi_{\theta_{\text{old}}}(o_i|q)} A_i, \, 
    \text{clip} \bigg( 
    \frac{\pi_\theta(o_i|q)}{\pi_{\theta_{\text{old}}}(o_i|q)}, \, 1 - \epsilon, \, 1 + \epsilon 
    \bigg) A_i 
    \bigg)
    \label{eq:min_block}
\end{align}

\begin{align}
    \mathcal{D}_{\text{KL}} \big( \pi_\theta \| \pi_{\text{ref}} \big) = 
    \frac{\pi_{\text{ref}}(o_i|q)}{\pi_\theta(o_i|q)} 
    - \log \frac{\pi_{\text{ref}}(o_i|q)}{\pi_\theta(o_i|q)} - 1
    \label{eq:kl_divergence}
\end{align}

where the $\pi_\theta$ is the policy being optimized, $\pi_{\theta_{old}}$ is the previous policy iteration
, $\pi_{ref}$ is a reference policy for regularization
, $G$ is the group size for advantage calculation
, $D_{KL}$ is the reference policy divergence penalty, $\lambda$ is a regularization hyperparameter.
For each context $q$, GRPO samples a group of $G$ outputs from the old policy, calculates their rewards, and normalizes these rewards using shift-and-scale normalization. This approach enables the model to learn relative preferences among outputs rather than absolute reward values. 


Next, we propose a multi-dimensional Reward (R) system for enhancing factual accuracy in GRPO-based reasoning. The Factual Reward boosts truthfulness via our fact-checker, while the Semantic Reward aligns outputs with references. The Format Reward ensures consistent structure, and the Length Reward promotes suitable detail. These guide models to produce accurate reasoning chains.

\vspace{1mm}
\noindent\textbf{(R1). Factual Enhancement Reward.} 
The factual enhancement reward evaluates the accuracy of each reasoning step using our specialized fact-checking model $\pi_\theta^*$:

\begin{equation}
R_{\text{fact}}(T, S|q) = \frac{1}{|T_{\text{valid}}|} \sum_{t_i \in T_{\text{valid}}} \mathbb{I}(P_{\pi_\theta^*}(t_i|q) > \tau)
\label{eq: fact-reward}
\end{equation}

where $P_{\pi_\theta^*}(t_i|q)$ represents the probability assigned by our fact-checking model that reasoning step $t_i$ is factually correct given query $q$, $\tau$ is a factuality threshold parameter, $\mathbb{I}(\cdot)$ is the indicator function, and $T_{\text{valid}} \subseteq T$ is the subset of reasoning steps that meet length requirements. If $T_{\text{valid}}$ is empty, $R_{\text{fact}}(T, S|q) = 0$. 
This formulation incorporates length verification by defining $T_{\text{valid}} = \{t_i \in T \mid L_{\text{min}} \leq L(t_i) \leq L_{\text{max}}\}$, where $L(t_i)$ is the token count of step $t_i$, and $[L_{\text{min}}, L_{\text{max}}]$ defines the acceptable range. This prevents the model from artificially inflating factuality scores by generating trivially short reasoning steps.

\vspace{1mm}
\noindent\textbf{(R2). Semantic Similarity Reward.}
The semantic similarity reward ensures that the generated final answer aligns closely with the reference solution (i.e., ground truth answer):

\begin{equation}
R_{\text{sim}}(T, S|q) = \text{sim}(E(S(q)), E(S^*(q)))
\label{eq: sim-reward}
\end{equation}

\noindent where $S^*$ represents the reference solution for query $q$, $E(\cdot)$ denotes the embedding function mapping text to a dense vector representation using a sentence transformer model, and $\text{sim}(\cdot,\cdot)$ computes the cosine similarity between these embeddings. This reward component ensures that even as the model optimizes for factual reasoning, the final answers remain semantically aligned with the expected solutions.  For implementation, we extract the final answer using template-specific markers (e.g., \textit{``boxed\{answer content\}''}) and compare it against similarly extracted content from the reference solution. The reward returns a positive value when the similarity exceeds a threshold $\delta$, and a negative value otherwise, creating a clear incentive for maintaining answer correctness.

\vspace{1mm}
\noindent\textbf{(R3). Format Compliance Reward.}
The format compliance reward encourages adherence to the expected structural conventions:

\begin{equation}
R_{\text{format}}(T, S|q) = 
\begin{cases}
\alpha & \text{if valid format is found} \\
-\beta & \text{otherwise}
\end{cases}
\label{eq: format-reward}
\end{equation}

where $\alpha > 0$ and $\beta > 0$ are positive constants. This reward verifies that the response contains exactly one properly formatted solution section with appropriate delineation markers, ensuring consistent and machine-interpretable outputs.
For instance, in reasoning tasks using structured templates, a valid response should contain precisely one reasoning section enclosed in appropriate tags (e.g., \textit{``$<$think$>$''} and \textit{``$<$/think$>$''}), followed by exactly one solution enclosed in standardized notation (e.g., \textit{boxed\{...\}}): \textit{``$<$think$>$To solve this problem, I need to... When... $<$/think$>$ boxed\{answer\}''}.

This format reward function guarantees consistent adherence to structural conventions across all model responses, facilitating automated evaluation and improving readability. The binary reward structure creates a clear incentive for maintaining proper formatting while penalizing responses that either omit required formatting elements or include redundant ones.

\vspace{1mm}
\noindent\textbf{(R4). Length Constraint Reward.}
The length Constraint reward promotes reasoning of suitable detail and conciseness:

\begin{equation}
R_{\text{length}}(T, S|q) =
\begin{cases}
\gamma & \text{if } L_{\text{min}}^{\text{total}} \leq L(T \oplus S) \leq L_{\text{max}}^{\text{total}} \\
-\eta & \text{otherwise}
\end{cases}
\label{eq: length-reward}
\end{equation}

where $L(T \oplus S)$ represents the total token count of the combined reasoning steps and solution, $L_{\text{min}}^{\text{total}}$ and $L_{\text{max}}^{\text{total}}$ define the acceptable bounds for response length, and $\gamma > 0$ and $\eta > 0$ are positive constants. These bounds are empirically determined to discourage both overly brief explanations that lack thoroughness and excessively verbose responses that might introduce errors or redundancies.



This multi-faceted reward provides a comprehensive assessment framework that guides our GRPO-based optimization toward generating reasoning chains that are factually correct, lead to accurate final answers, maintain proper structure, and provide appropriate detail.

\subsection{Reasoning Trajectory Analysis (RQ3)}
\label{subsec:interpretability}

To understand how factuality enhancement training influences the internal reasoning process steps of LLMs, we must move beyond output-level evaluations and examine how such training reshapes the model's neural activation dynamics during multi-step reasoning. 

Our central hypothesis is that improvements in factuality manifest as measurable changes in the trajectory of hidden state activations across reasoning steps, such as smoother progression, increased coherence, or more stable directionality. By quantifying these properties, we can gain mechanistic insights into how factual reasoning emerges within the model. More importantly, understanding these internal patterns offers a foundation for developing future training strategies: if certain activation signatures are reliably associated with factual correctness, they can serve as targets or regularization signals to guide more factually robust model training. 

\vspace{1mm}
\noindent{\textbf{Reasoning Trajectory Activation Definition}}
 {To analyze how factuality enhancement training reshapes the reasoning LLM's internal activation dynamics, we quantitatively trace the evolution of hidden state neural activations across reasoning steps. This approach enables us to inspect the internal transformations that occur during multi-step reasoning and understand the mechanistic changes induced by \MethodName.}
We formalize the analysis of model reasoning steps through \textbf{latent activations} (hidden state activations). For a language model with $L$ transformer layers processing a reasoning sequence with $T$ steps: the $A_{l,t}$ represents the mean hidden state activation at layer $l \in \{1,...,L\}$ for reasoning step $t \in \{1,...,T\}$.
We compute the activations as:

\begin{equation}
A_{l,t} = \frac{1}{|S_t|} \sum_{i \in S_t} h_{l,i}
\label{eq:activations}
\end{equation}

where $S_t$ is the set of token positions in step $t$, and $h_{l,i}$ is the hidden state at layer $l$ and token position $i$.

\vspace{1mm}
\noindent{\textbf{Trajectory Characterization Metrics}}
To characterize the trajectory properties of our model, we employ several key metrics. First, we measure the Euclidean distance between consecutive reasoning steps at layer $l$, defined as:

\begin{equation}
D_l(t, t+1) = \|A_{l,t+1} - A_{l,t}\|_2
\label{eq:distance}
\end{equation}

where $\|\cdot\|_2$ denotes the Euclidean norm, quantifying the magnitude of change in representational space between steps $t$ and $t+1$ at layer $l$.
To quantify the overall reasoning divergence throughout the entire trajectory, we calculate the mean step distance as:

\begin{equation}
\bar{D}_l = \frac{1}{T-1} \sum_{t=1}^{T-1} D_l(t, t+1)
\label{eq:distance2}
\end{equation}

where $\bar{D}_l$ represents the average distance between consecutive reasoning steps at layer $l$, and $T$ is the total number of reasoning steps. This aggregate measure helps us understand how significantly the intermediate representations change during the reasoning process, with larger values indicating more substantial transformations between consecutive steps.

\vspace{1mm}
\noindent{\textbf{Angular Deviation}}
We calculate the angular change in reasoning steps direction:

\begin{equation}
\theta_l(t) = \arccos\left(\frac{(A_{l,t+1} - A_{l,t}) \cdot (A_{l,t+2} - A_{l,t+1})}{\|A_{l,t+1} - A_{l,t}\|_2 \cdot \|A_{l,t+2} - A_{l,t+1}\|_2}\right)
\label{eq:angular}
\end{equation}

where $\theta_l(t)$ represents the angle between consecutive reasoning steps at time steps $t$, $t+1$, and $t+2$ for layer $l$. The numerator computes the dot product between adjacent step latent activations, while the denominator normalizes by their respective magnitudes. The $\arccos$ function converts this normalized dot product to an angle in radians. The mean angular deviation across steps:

\begin{equation}
\bar{\theta}_l = \frac{1}{T-1} \sum_{t=1}^{T-1} \theta_l(t)
\label{eq:angular2}
\end{equation}

where $\bar{\theta}_l$ denotes the average angular deviation across all reasoning steps at layer $l$, and $T$ is the total number of reasoning steps. The summation is intentionally limited to the first $T-1$ steps to exclude the final answer portion of the LLM response, thereby focusing exclusively on the intermediate reasoning process. Higher values indicate more exploratory reasoning patterns with frequent directional changes, while lower values suggest more consistent, directed reasoning trajectories.

\vspace{1mm}
\noindent{\textbf{Step Coherence Matrix}}
We compute the cosine similarity between all pairs of reasoning steps:

\begin{equation}
C_l(i,j) = \frac{A_{l,i} \cdot A_{l,j}}{\|A_{l,i}\|_2 \cdot \|A_{l,j}\|_2}
\end{equation}

where $C_l(i,j)$ represents the cosine similarity between latent activations at reasoning steps $i$ and $j$ for layer $l$. This metric quantifies the alignment of representational directions, with values ranging from -1 (completely opposite) to 1 (perfectly aligned).
Adjacent step similarity measures local consistency:

\begin{equation}
S_l = \frac{1}{T-1} \sum_{t=1}^{T-1} C_l(t, t+1)
\end{equation}

where $S_l$ denotes the average cosine similarity between consecutive reasoning steps at layer $l$, providing a measure of local coherence throughout the reasoning steps. Higher values of $S_l$ indicate more consistent progression between adjacent reasoning steps.



To visualize high-dimensional trajectories, we employ the dimensionality reduction technique. We utilize PCA to project activations to a 2D space that maximizes variance: $Z^{PCA}_{l,t} = \text{PCA}(\{A_{l,t}\}_{t=1}^T)$ represents the 2D projection of the original activation $A_{l,t}$ using PCA. The explained variance ratio $\lambda_i$ for each principal component $i$ provides confidence in the projection fidelity, allowing us to assess how well the 2D representation captures the original high-dimensional dynamics.


\section{Experiments Setup}

\begin{table*}[t]
\centering
\begin{threeparttable}
\scriptsize 
\caption{Comparison of Fact-Checking Performance Across Different Models and Training Approaches (Prompt Engineering, Chain-of-Thought, and LoRA Fine-tuning).}
\label{tab:model_classifier_comparison}
\setlength{\tabcolsep}{2.5pt}  
\begin{tabular}{l|cccc|cccc||cccc|cccc||cccc|cccc}
\toprule
& \multicolumn{8}{c||}{\textbf{Prompt Engineering~\cite{DBLP:journals/corr/abs-2402-07927}}} & \multicolumn{8}{c||}{\textbf{Chain-of-Thought~\cite{DBLP:conf/nips/Wei0SBIXCLZ22}}} & \multicolumn{8}{c}{\textbf{LoRA Fine-tuning~\cite{DBLP:conf/iclr/HuSWALWWC22}}} \\
\midrule
\textbf{Model} & \textbf{TP} & \textbf{FP} & \textbf{FN} & \textbf{TN} & \textbf{Acc.$\uparrow$} & \textbf{Prec.$\uparrow$} & \textbf{Rec.$\uparrow$} & \textbf{F1.$\uparrow$} & 
\textbf{TP} & \textbf{FP} & \textbf{FN} & \textbf{TN} & \textbf{Acc.$\uparrow$} & \textbf{Prec.$\uparrow$} & \textbf{Rec.$\uparrow$} & \textbf{F1.$\uparrow$} & 
\textbf{TP} & \textbf{FP} & \textbf{FN} & \textbf{TN} & \textbf{Acc.$\uparrow$} & \textbf{Prec.$\uparrow$} & \textbf{Rec.$\uparrow$} & \textbf{F1.$\uparrow$} \\
\midrule
Llama-3.2-1B-Ins & 61 & 406 & 70 & 463 & 52.40 & 13.06 & 46.56 & 0.20 & 94 & 661 & 37 & 208 & 30.20 & 12.54 & 71.76 & 0.21 & 103 & 23 & 28 & 846 & 94.90 & 81.75 & 78.63 & 0.80 \\
Llama-3.2-3B-Ins & 92 & 527 & 39 & 342 & 43.40 & 14.86 & 70.23 & 0.25 & 93 & 550 & 38 & 319 & 41.20 & 14.46 & 70.99 & 0.24 & \cellcolor{gray!20}116 & \cellcolor{gray!20}13 & \cellcolor{gray!20}15 & \cellcolor{gray!20}856 & \cellcolor{gray!20}\textbf{97.20} & \cellcolor{gray!20}89.92 & \cellcolor{gray!20}88.55 & \cellcolor{gray!20}\textbf{0.89} \\
Qwen2.5-0.5B-Ins & 112 & 774 & 19 & 95 & 20.70 & 12.64 & 85.50 & 0.22 & 80 & 503 & 51 & 366 & 44.60 & 13.72 & 61.07 & 0.22 & 92 & 28 & 40 & 848 & 93.25 & 76.67 & 69.70 & 0.73 \\
Qwen2.5-1.5B-Ins & 18 & 132 & 113 & 737 & 75.50 & 12.00 & 13.74 & 0.13 & 120 & 791 & 11 & 78 & 19.80 & 13.17 & 91.60 & 0.23 & 104 & 22 & 28 & 854 & 95.04 & 82.54 & 78.79 & 0.81 \\
Qwen2.5-7B-Ins & \cellcolor{gray!20}80 & \cellcolor{gray!20}366 & \cellcolor{gray!20}51 & \cellcolor{gray!20}503 & \cellcolor{gray!20}58.30 & \cellcolor{gray!20}17.94 & \cellcolor{gray!20}61.07 & \cellcolor{gray!20}0.28 & \cellcolor{gray!20}58 & \cellcolor{gray!20}322 & \cellcolor{gray!20}73 & \cellcolor{gray!20}547 & \cellcolor{gray!20}\textbf{60.50} & \cellcolor{gray!20}15.26 & \cellcolor{gray!20}44.27 & \cellcolor{gray!20}0.23 & 116 & 20 & 15 & 849 & 96.50 & 85.29 & 88.55 & 0.87 \\
\bottomrule
\end{tabular}
\begin{tablenotes}
\footnotesize
\item[*] Ins = Instruct; Acc, Prec, Rec = (\%)
\end{tablenotes}
\end{threeparttable}
\end{table*}

\subsection{Dataset}

\noindent{\textbf{Fact-Checking Dataset:}}
For the LLM's factuality checking task, we utilized the converted chain-of-thought data described in previous section §~\ref{subsec:fact-checking}. By replacing facts and entities through Named Entity Recognition (NER), we collected over 38,539 data points. This dataset serves as the foundation for training our fact detection classifier. We selected 1,000 instances for our validation set and another 1,000 for our test set. The model's classification performance on this test set serves as the definitive measure of our fact detection model's effectiveness. We deliberately deviated from the conventional 80-10-10 split used in traditional machine learning because autoregressive generative models, with their enormous parameter counts, typically train on datasets vastly larger than their test sets.

\vspace{1mm}
\noindent{\textbf{Factuality Enhancement Dataset:}}
For the factuality enhancement task, we generated appropriate questions from the original Wiki data by calling the \textit{DeepSeek-671B}~\cite{DBLP:journals/corr/abs-2501-12948} API, and then produced usable chain-of-thought reasoning data using the \textit{DeepSeek-671B} model. We employed LLM-as-a-judge~\cite{DBLP:journals/corr/abs-2411-15594} (GPT-4o) to select and filter the final 6,000 reasoning samples used to enhance the model's factuality. We designated 100 instances for our validation set and selected 100 instances for our final test set to evaluate model performance.

\subsection{Models}
For the comprehensive evaluation, we employ a diverse set of open-source and commercial models across three interconnected tasks. For the fact-checking classifier training, we utilize two model families: \textit{Llama-3.2-Instruct} \cite{DBLP:journals/corr/abs-2302-13971} (1B and 3B parameters) and \textit{Qwen2.5-Instruct} \cite{qwen} (0.5B, 1.5B, and 7B parameters), selecting these instruction-tuned variants for their directive-following capabilities and to demonstrate cross-architecture generalizability. To evaluate factuality within reasoning processes, we tested a broader spectrum of models including specialized \textit{Qwen2.5-0.5B} variants (Distill and GRPO-trained) \cite{openr1}, \textit{DeepSeek-R1-Distill-Qwen} models (1.5B and 7B) \cite{DBLP:journals/corr/abs-2501-12948}, larger-scale models (Qwen \textit{QwQ-32B} \cite{qwq32b} and \textit{DeepSeek-671B} \cite{DBLP:journals/corr/abs-2501-12948}), and commercial black-box models (\textit{GPT-o1}~\cite{OpenAIo1}, \textit{Gemini2.0-Flash-Thinking}~\cite{Gemini2}, and \textit{Claude-3.7-Sonnet-Thinking}~\cite{Claude-3.7}), with evaluations conducted across temperature settings from 0.3 to 1.0. 

For the reasoning process factual enhancement task, we focused on four white-box models allowing parameter access and modification: \textit{Qwen2.5-0.5B-Open-R1-Distill}, its math-enhanced variant \textit{Qwen2.5-0.5B-Open-R1-Distill-Math-GRPO}, and two \textit{DeepSeek-R1-Distill-Qwen models} (1.5B and 7B parameters, distilled from a 671B model), enabling analysis of how architectural differences and parameter scales (ranging from 0.5B to 7B) affect the effectiveness of reasoning correction.

\subsection{Experiment Environment}
All our experiments were conducted on \textit{Ubuntu 22.04}. For the fact-checking models training, we utilized the \textit{LLama-Factory}~\cite{DBLP:journals/corr/abs-2403-13372} framework, which builds upon the \textit{Transformers} library~\cite{DBLP:journals/corr/abs-1910-03771} and \textit{PyTorch}. For our fact enhancement algorithms using reinforcement learning, we employed the \textit{Transformer Reinforcement Learning (TRL)} package as the foundation.
For the final answer quality evaluations, we use the \textit{LightEval}~\cite{lighteval} package from huggingface.
For GRPO fine-tuning, we used the \textit{Open-R1}~\cite{openr1} framework as our experimental base and extended it with our custom-designed fact enhancement reward strategies. To enable efficient sample generation during training, we employ the \textit{vLLM}~\cite{kwon2023efficient} library.
Our experiments were conducted on machines equipped with two H20-96G GPUs and four RTX 3090-24G GPUs. All fine-tuning experiments were performed on the four H20-96G GPUs, while mechanical interpretability analysis was performed on four RTX 3090-24G GPUs machine. To address the substantial GPU resource requirements for fine-tuning, we implemented \textit{deepspeed}~\cite{DBLP:journals/corr/abs-2207-00032} technology for distributed multi-GPUs training.

\subsection{Metrics}

\noindent{\textbf{Reasoning Process Fact-Checking Classifier:}}
For evaluating our fact-checking classifier, we employ a binary classification framework with accuracy as our primary metric. Accuracy measures the proportion of correctly classified instances (both factual and non-factual reasoning steps). We also report \textbf{Precision}, \textbf{Recall}, and \textbf{F1} score to provide a more comprehensive assessment, particularly important given potential class imbalances. Our implementation enforces strict output formatting, where predictions must exactly match the expected format (e.g., \textit{``$<$fact$>$True$<$/fact$>$''}) to ensure standardized outputs for downstream applications. 





\vspace{1mm}
\noindent{\textbf{Reasoning Process Enhancement Training:}}
For evaluating factual enhancement training, we segment reasoning chains into discrete steps and assess factuality at each step. Given a reasoning chain $T$ produced for question $q$, we segment $T$ into a sequence of discrete steps $T = \{t_1, t_2, ..., t_n\}$. Our primary metric is Reasoning Chain Steps Factuality Accuracy (SFA), defined as: $\text{Acc}(T) = \frac{1}{n}\sum_{i=1}^{n} f_\theta(q, t_i)$ where $f_\theta(q, t_i)$ evaluates whether step segment $t_i$ is factually consistent given question $q$. To assess robustness to generation parameters, we evaluate reasoning steps across multiple temperature settings and measure variance in factuality performance: $\text{Variance (Var.)} = \sigma^2({\text{Acc}(T) | t \in T})$ where $T$ represents the reasoning chain generated at specific temperature. Lower variance values indicate more consistent factual reasoning across different sampling conditions.
\section{Experiments Results}

\begin{table*}[t]
\centering
\scriptsize   
\setlength{\tabcolsep}{1.5pt}  
\setlength{\arrayrulewidth}{0.3pt}  
\caption{Performance comparison of different models. Accuracy ($\uparrow$) indicates higher is better, Variance ($\downarrow$) indicates lower is better. For each model, we present performance across different temperature settings.}
\label{tab:llms_models_fact_comparison}
\begin{tabular}{l|c|c|*{14}{c}}
\toprule
\multirow{3}{*}{\textbf{Model}} & \multirow{3}{*}{\textbf{\begin{tabular}[c]{@{}c@{}}Overall\\Acc. (\%)$\uparrow$\end{tabular}}} & \multirow{3}{*}{\textbf{\begin{tabular}[c]{@{}c@{}}Overall\\Var.$\downarrow$\end{tabular}}} & \multicolumn{14}{c}{\textbf{Temperature}} \\
\cmidrule{4-17}
 &  &  & \multicolumn{2}{c}{\textbf{0.3}} & \multicolumn{2}{c}{\textbf{0.4}} & \multicolumn{2}{c}{\textbf{0.5}} & \multicolumn{2}{c}{\textbf{0.6}} & \multicolumn{2}{c}{\textbf{0.7}} & \multicolumn{2}{c}{\textbf{0.8}} & \multicolumn{2}{c}{\textbf{1.0}} \\
\cmidrule{4-17}
 &  &  & \textbf{Acc.(\%)$\uparrow$} & \textbf{Var.$\downarrow$} & \textbf{Acc.(\%)$\uparrow$} & \textbf{Var.$\downarrow$} & \textbf{Acc.(\%)$\uparrow$} & \textbf{Var.$\downarrow$} & \textbf{Acc.(\%)$\uparrow$} & \textbf{Var.$\downarrow$} & \textbf{Acc.(\%)$\uparrow$} & \textbf{Var.$\downarrow$} & \textbf{Acc.(\%)$\uparrow$} & \textbf{Var.$\downarrow$} & \textbf{Acc.(\%)$\uparrow$} & \textbf{Var.$\downarrow$} \\
\midrule
\multicolumn{17}{c}{\textbf{White-box Models}} \\
\midrule
DeepSeek-R1-Distill-Qwen-1.5B & 68.06 & 0.218 & 72.54 & 0.199 & 69.57 & 0.212 & 70.63 & 0.208 & 68.57 & 0.216 & 63.49 & 0.232 & 63.73 & 0.231 & 62.31 & 0.236 \\
DeepSeek-R1-Distill-Qwen-7B & 71.46 & 0.151 & 71.24 & 0.152 & 72.57 & 0.144 & 73.93 & 0.137 & 72.12 & 0.145 & 70.37 & 0.157 & 78.56 & 0.169 & 77.10 & 0.174  \\
Qwen2.5-0.5B-Open-R1-GRPO & 73.66 & 0.194 & 76.28 & 0.181 & 73.16 & 0.197 & 74.99 & 0.188 & 74.91 & 0.188 & 71.92 & 0.202 & 70.72 & 0.207 & 68.53 & 0.213 \\
Qwen2.5-0.5B-Open-R1-Distill & 42.24 & 0.244 & 59.12 & 0.242 & 60.22 & 0.240 & 55.66 & 0.247 & 55.16 & 0.247 & 49.49 & 0.250 & 38.83 & 0.237 & 24.57 & 0.180 \\
Qwen QwQ-32B & 66.01 & 0.224 & 70.11 & 0.209 & 70.20 & 0.208 & 70.77 & 0.207 & 66.18 & 0.224 & 65.44 & 0.226 & 64.54 & 0.228 & 62.43 & 0.234 \\
DeepSeek-671B & 76.85 & 0.143 & 77.62 & 0.139 & 78.14 & 0.137 & 77.88 & 0.138 & 77.36 & 0.140 & 76.85 & 0.143 & 75.94 & 0.147 & 74.28 & 0.152 \\
\midrule
\multicolumn{17}{c}{\textbf{Black-box Models}} \\
\midrule
GPT-o1 & 82.57 & 0.148 & 84.33 & 0.135 & 83.95 & 0.138 & 83.21 & 0.142 & 82.76 & 0.145 & 82.19 & 0.149 & 81.58 & 0.155 & 80.72 & 0.162 \\
GPT-o1-mini & 74.05 & 0.192 & 73.66 & 0.194 & 73.18 & 0.197 & 72.88 & 0.200 & 75.86 & 0.183 & 74.92 & 0.186 & 73.84 & 0.192 & 72.62 & 0.198 \\
Gemini-2.0-flash-thinking & 83.00 & 0.141 & 84.22 & 0.132 & 83.91 & 0.134 & 83.58 & 0.137 & 83.17 & 0.139 & 82.75 & 0.143 & 82.11 & 0.147 & 81.46 & 0.150\\
Claude-3.7-sonnet-thinking & 81.93 & 0.179 & 76.53 & 0.173 & 78.65 & 0.171 & 79.84 & 0.170 & 83.77 & 0.167 & 81.22 & 0.169 & 80.45 & 0.172 & 79.49 & 0.168 \\
\bottomrule
\end{tabular}

\end{table*}

\subsection{Fact-Checking Evaluation (RQ1)}
\label{subsec:fact-checking-performance}

\noindent\textbf{Comparison of Different Methods for Fact-Checking:}
Our experimental results demonstrate that performance varies significantly across different approaches to fact-checking classification, with LoRA fine-tuning consistently outperforming both prompt engineering~\cite{DBLP:journals/corr/abs-2402-07927} and chain-of-thought~\cite{DBLP:conf/nips/Wei0SBIXCLZ22} methods. As shown in~\autoref{tab:model_classifier_comparison}, the \textit{Llama-3.2-3B-Instruct} model fine-tuned with LoRA achieved the best performance, with an accuracy of 97.20\% and an F1 score of 0.89. This model exhibited a well-balanced precision-recall trade-off (89.92\% precision, 88.55\% recall), indicating its strong capability to identify both factually correct and incorrect reasoning chain steps without significant bias toward either class. The \textit{Qwen2.5-7B-Instruct} model followed closely with 96.50\% accuracy and 0.87 F1 score when using LoRA fine-tuning.

Comparing across different approaches, we observe that prompt engineering and chain-of-thought methods perform substantially worse than LoRA fine-tuning for all tested models. The best performance from non-fine-tuning approaches was achieved by \textit{Qwen2.5-7B-Instruct} using chain-of-thought, reaching only 60.50\% accuracy and 0.23 F1 score. This stark performance gap (97.20\% vs. 60.50\% accuracy) demonstrates the necessity of parameter-efficient fine-tuning for effective fact-checking capabilities. Additionally, we note that model size within the same family correlates positively with fact-checking performance when using LoRA fine-tuning, as evidenced by the progression from \textit{Qwen2.5-0.5B} (F1: 0.73) to \textit{Qwen2.5-1.5B} (F1: 0.81) to \textit{Qwen2.5-7B} (F1: 0.87).

The extremely low F1 scores across prompt engineering (0.13-0.28) and chain-of-thought (0.21-0.24) methods suggest these approaches struggle with balancing precision and recall for fact-checking, often exhibiting high false positive or false negative rates. Analysis of failure cases reveals these methods misclassify ambiguous statements and struggle with complex negations. These findings align with prior research showing that LoRA-based fine-tuning better preserves model capabilities while enabling adaptation to specialized tasks like fact-checking, where maintaining a balanced understanding of factuality is critical~\cite{DBLP:journals/corr/abs-2405-09673}. Parameter-efficient fine-tuning appears superior due to its ability to modify relevant model weights while retaining pre-trained knowledge.

\vspace{2mm}
\findbox{
Our work finds that LoRA fine-tuning far outperforms prompt engineering and CoT methods for reasoning steps fact-checking classification. While prompt and CoT approaches struggle with low F1 scores and poor precision-recall balance, LoRA-tuned models achieve high accuracy and robust performance. The results highlight that parameter-efficient fine-tuning is essential for reliable fact-checking.}

\vspace{2mm}
\noindent\textbf{Comparison of Different LLMs for Fact-Checking:}
Our comprehensive evaluation of reasoning factuality across various LLMs reveals significant variations in performance based on model architecture, size, and generation parameters. ~\autoref{tab:llms_models_fact_comparison} presents our findings across different temperature settings for both white-box and black-box models. 
Among white-box models, \textit{DeepSeek-671B} achieves the highest factual accuracy at 76.85\%. This aligns with expectations that larger models typically demonstrate superior reasoning capabilities. Interestingly, \textit{Qwen2.5-0.5B-Open-R1-GRPO} shows strong performance with 73.66\% accuracy despite its smaller size, followed by \textit{DeepSeek-R1-Distill-Qwen-7B} at 71.46\%. 
The substantially larger \textit{QwQ-32B} model achieves only 66.01\% accuracy, underperforming relative to its scale. 
Specialized training like GRPO can boost factual reasoning in smaller models, suggesting efficiency over model scaling.

Black-box models consistently outperform their white-box counterparts, with \textit{Gemini-2.0-flash-thinking} achieving the highest overall factual accuracy at 83.00\%, closely followed by \textit{GPT-o1} at 82.57\% and \textit{Claude-3.7-sonnet-thinking} at 81.93\%. 
This substantial performance gap (over 6 percentage points between the best black-box and white-box models) indicates that proprietary models maintain significant advantages in factual reasoning. Our temperature analysis reveals a critical finding: virtually all models exhibit sensitivity to temperature settings, with performance generally declining as temperature increases beyond optimal ranges. 
For instance, \textit{DeepSeek-R1-Distill-Qwen-7B} shows significant variance across temperature settings, with peak performance at 0.5 (73.93\%) dropping to 70.37\% at 0.7, before exhibiting an unexpected increase at higher temperatures. Most models, including black-box models like \textit{GPT-o1}, demonstrate their best performance at lower temperature settings (0.3-0.5), with gradual degradation as temperatures increase. The exception is \textit{Claude-3.7-sonnet-thinking}, which shows atypical behavior with its lowest accuracy (76.53\%) at temperature 0.3 and peak performance (83.77\%) at temperature 0.6.

\vspace{2mm}
\findbox{
These results demonstrate that current mainstream reasoning LLMs consistently exhibit factual errors during reasoning. This raises significant concerns regarding the reliability of such models in reasoning-intensive applications, particularly for white-box models that remain more accessible but less capable than their proprietary counterparts.\looseness=-1 }

\subsection{Factuality Enhancement Evaluation (RQ2)}
\label{subsec:fact-enhancement-res}


We use open-source models for fact-enhanced training and assess how reasoning steps improve factual accuracy. We evaluate gains and analyze if reasoning improvements affect response quality.

\begin{table*}[t]
\centering
\scriptsize 
\setlength{\tabcolsep}{2.5pt}  
\setlength{\arrayrulewidth}{0.3pt}  
\caption{Comparison of Model Performance Across Different Temperatures. Average of temperatures 0.9 and 1.0 for Model 2 base version. Best results for each model are shown in \textbf{bold}. Shaded cells indicate relative improvement after enhancement. ↑ indicates higher values are better, ↓ indicates lower values are better. Variance reduction indicates more consistent performance across samples.}
\label{tab:fact-grpo-improve}
\begin{tabular}{c|ccccccccccccccc|cc}
\toprule
\multirow{3}{*}{\textbf{Model}} & \multicolumn{15}{c|}{\textbf{Performance at Different Temperatures}} & \multicolumn{2}{c}{\textbf{Overall}} \\
\cmidrule(lr){2-16} \cmidrule(lr){17-18}
& \multicolumn{2}{c}{\textbf{0.3}} & \multicolumn{2}{c}{\textbf{0.4}} & \multicolumn{2}{c}{\textbf{0.5}} & \multicolumn{2}{c}{\textbf{0.6}} & \multicolumn{2}{c}{\textbf{0.7}} & \multicolumn{2}{c}{\textbf{0.8}} & \multicolumn{2}{c}{\textbf{1.0}} & \multicolumn{1}{c|}{} & \textbf{Acc.(\%)} & \textbf{Var.} \\
& \textbf{Acc.(\%)$\uparrow$} & \textbf{Var.$\downarrow$} & \textbf{Acc.(\%)$\uparrow$} & \textbf{Var.$\downarrow$} & \textbf{Acc.(\%)$\uparrow$} & \textbf{Var.$\downarrow$} & \textbf{Acc.(\%)$\uparrow$} & \textbf{Var.$\downarrow$} & \textbf{Acc.(\%)$\uparrow$} & \textbf{Var.$\downarrow$} & \textbf{Acc.(\%)$\uparrow$} & \textbf{Var.$\downarrow$} & \textbf{Acc.(\%)$\uparrow$} & \textbf{Var.$\downarrow$} & \multicolumn{1}{c|}{} & \textbf{↑} & \textbf{↓} \\

\midrule
\multicolumn{18}{c}{\textit{DeepSeek-R1-Distill-Qwen-1.5B}} \\
\midrule
BaseModel & 72.55 & 0.199 & 69.56 & 0.212 & 70.59 & 0.208 & 68.55 & 0.216 & 63.31 & 0.232 & 63.73 & 0.231 & 59.89 & 0.238 & \multicolumn{1}{c|}{} & 68.53 & 0.217 \\
+ RELIANCE & \textbf{82.22} & \textbf{0.146} & \textbf{86.67} & \textbf{0.116} & \textbf{83.10} & \textbf{0.159} & \textbf{81.10} & \textbf{0.153} & \textbf{84.89} & \textbf{0.129} & \textbf{80.54} & \textbf{0.157} & \textbf{79.42} & \textbf{0.162} & \multicolumn{1}{c|}{} & \textbf{82.65} & \textbf{0.143} \\
\rowcolor{gray!20}
\textit{Improvement} & +9.70 & -0.053 & +17.10 & -0.096 & +9.60 & -0.049 & +12.60 & -0.063 & +21.50 & -0.103 & +16.80 & -0.074 & +19.60 & -0.076 & \multicolumn{1}{c|}{} & +14.12 & -0.074 \\
\midrule
\multicolumn{18}{c}{\textit{DeepSeek-R1-Distill-Qwen-7B}} \\
\midrule
BaseModel & 71.26 & 0.152 & 72.57 & 0.144 & 73.95 & 0.137 & 72.15 & 0.145 & 70.31 & 0.157 & 78.59 & 0.169 & 68.74 & 0.175 & \multicolumn{1}{c|}{} & 71.40 & 0.151 \\

RELIANCE & \textbf{79.49} & \textbf{0.099} & \textbf{82.51} & \textbf{0.070} & \textbf{81.51} & \textbf{0.082} & \textbf{80.78} & \textbf{0.087} & \textbf{78.94} & \textbf{0.104} & \textbf{77.53} & \textbf{0.112} & \textbf{77.27} & \textbf{0.118} & \multicolumn{1}{c|}{} & \textbf{80.17} & \textbf{0.092} \\
\rowcolor{gray!20}
\textit{Improvement} & +8.20 & -0.053 & +10.00 & -0.074 & +7.60 & -0.055 & +8.60 & -0.058 & +8.60 & -0.053 & +9.00 & -0.057 & +8.50 & -0.057 & \multicolumn{1}{c|}{} & +8.77 & -0.059 \\

\midrule

\multicolumn{18}{c}{\textit{Qwen2.5-0.5B-Open-R1-Distill}} \\

\midrule
BaseModel & 59.15 & 0.242 & 60.28 & 0.240 & 55.69 & 0.247 & 55.12 & 0.247 & 49.48 & 0.250 & 38.81 & 0.237 & 24.53$^\dagger$ & 0.244 & \multicolumn{1}{c|}{} & 42.20 & 0.244 \\
+ RELIANCE & \textbf{92.47} & \textbf{0.070} & \textbf{92.67} & \textbf{0.068} & \textbf{91.84} & \textbf{0.075} & \textbf{92.19} & \textbf{0.073} & \textbf{91.82} & \textbf{0.075} & \textbf{91.91} & \textbf{0.074} & \textbf{90.86} & \textbf{0.082} & \multicolumn{1}{c|}{} & \textbf{92.10} & \textbf{0.073} \\
\rowcolor{gray!20}
\textit{Improvement} & +33.30 & -0.172 & +32.40 & -0.172 & +36.20 & -0.172 & +37.00 & -0.174 & +42.40 & -0.175 & +53.10 & -0.163 & +66.30 & -0.162 & \multicolumn{1}{c|}{} & +49.90 & -0.171 \\
\midrule
\multicolumn{18}{c}{\textit{Qwen2.5-0.5B-Open-R1-Distill-Math-GRPO}} \\
\midrule
BaseModel & 76.27 & 0.181 & 73.12 & 0.197 & 74.93 & 0.188 & 74.99 & 0.188 & 71.91 & 0.202 & 70.72 & 0.207 & 65.36 & 0.218 & \multicolumn{1}{c|}{} & 73.60 & 0.194 \\
+ RELIANCE & \textbf{88.00} & \textbf{0.106} & \textbf{89.90} & \textbf{0.091} & \textbf{89.50} & \textbf{0.094} & \textbf{85.30} & \textbf{0.125} & \textbf{85.20} & \textbf{0.126} & \textbf{87.20} & \textbf{0.112} & \textbf{84.50} & \textbf{0.128} & \multicolumn{1}{c|}{} & \textbf{87.50} & \textbf{0.109} \\
\rowcolor{gray!20}
\textit{Improvement} & +11.80 & -0.075 & +17.10 & -0.106 & +9.60 & -0.094 & +12.60 & -0.063 & +21.50 & -0.076 & +16.80 & -0.071 & +19.20 & -0.090 & \multicolumn{1}{c|}{} & +14.60 & -0.074 \\

\bottomrule
\end{tabular}
\vspace{-3mm}
\end{table*}

\vspace{1mm}
\noindent\textbf{Factual Reasoning Accuracy: }
Our experimental results demonstrate that \MethodNameNoSpace significantly enhances factual accuracy in reasoning across all tested models, with particularly dramatic improvements in smaller models. As shown in ~\autoref{tab:fact-grpo-improve}, our approach consistently delivers substantial gains in factual reasoning accuracy while simultaneously reducing variance across different temperature settings. The \textit{Qwen2.5-0.5B-Open-R1-Distill} model exhibited the most remarkable improvement, with factual accuracy increasing from 42.20\% to 92.10\% (+49.90 percentage points) and variance reducing from 0.244 to 0.073 (-0.171), underscoring the effectiveness of our approach for resource-constrained models. This dramatic improvement suggests that \MethodNameNoSpace effectively addresses fundamental limitations in smaller models' factual reasoning capabilities, making them viable alternatives to much larger models for applications requiring reliable reasoning.

For larger models, we observed consistent but more moderate improvements. \textit{DeepSeek-R1-Distill-Qwen-7B} shows an 8.77 percentage point increase in overall accuracy (from 71.40\% to 80.17\%), while \textit{DeepSeek-R1-Distill-Qwen-1.5B} improved by 14.12 percentage points (from 68.53\% to 82.65\%). Notably, \MethodNameNoSpace not only improved average performance but also substantially enhanced reasoning stability, with all models showing significantly reduced variance across temperature settings. This stability improvement is particularly evident in the \textit{Qwen2.5-0.5B-Open-R1-Distill-Math-GRPO} model, where our approach increased accuracy from 73.60\% to 87.50\% while reducing variance by 0.085 (from 0.194 to 0.109), demonstrating that \MethodNameNoSpace effectively complements existing mathematical reasoning capabilities with enhanced factual robustness. These results confirm that our approach successfully addresses the critical challenge of maintaining factual consistency throughout multi-step reasoning processes even under varying inference conditions.




\begin{figure}[h]
  \centering
  \includegraphics[trim=0 0 0 0, clip, width=0.88\columnwidth]{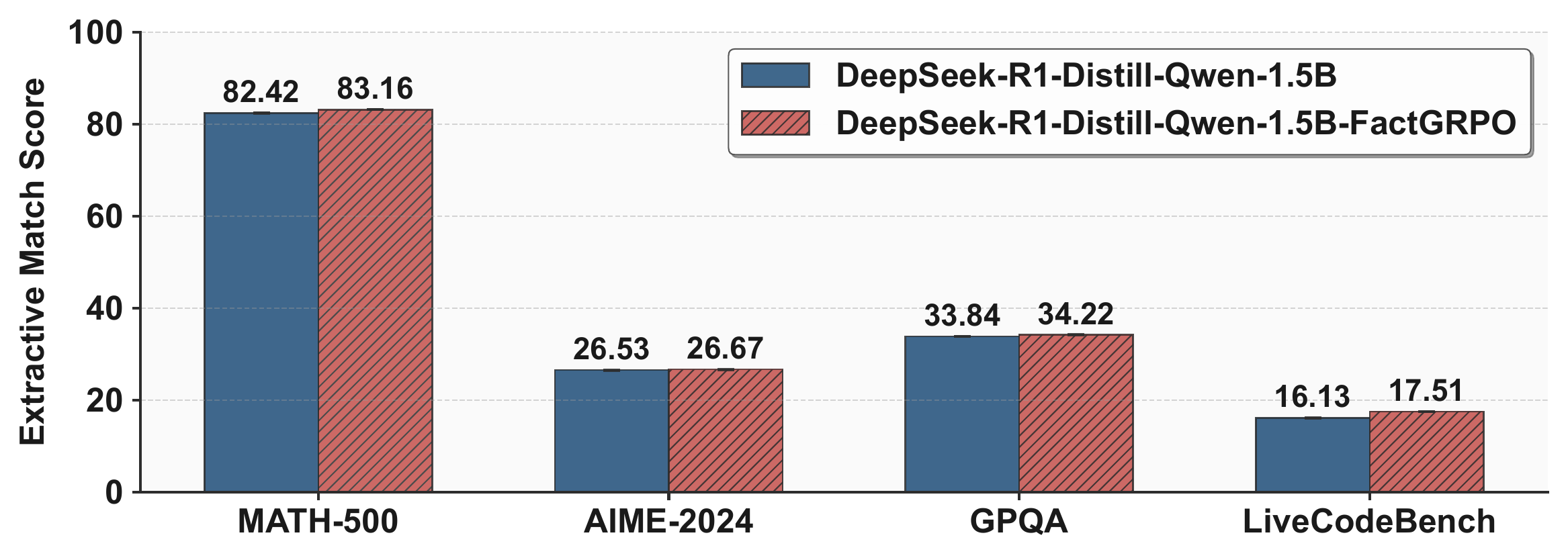}
  \caption{Extractive match accuracy (\%) comparison with and without \MethodNameNoSpace across diverse benchmarks.}
  \label{fig:benchmark_res}
  \vspace{-2mm}
\end{figure}

\vspace{1mm}
\noindent\textbf{Factual Reliability:}
While our \MethodNameNoSpace framework substantially enhances factual reasoning steps across models, a natural question arises: 
does this reasoning steps factuality improvement compromise the quality of the final reasoning output?
To address this, we conducted a comprehensive evaluation across several benchmark datasets. Our analysis reveals that \MethodNameNoSpace not only maintains task performance but can also yield modest improvements, while significantly reducing factual errors. As shown in~\autoref{fig:benchmark_res}, the \textit{DeepSeek-R1-Distill-Qwen-1.5B} model augmented with \MethodNameNoSpace achieves slightly higher accuracy on Math-500 (83.16\% vs. 82.42\%) and AIME-2024 (26.67\% vs. 26.53\%) compared to the baseline. Performance remains largely consistent on GPQA (34.22\% vs. 33.84\%) and LiveCodeBench (17.51\% vs. 16.13\%). 
\vspace{2mm}
\findbox{The \MethodNameNoSpace framework significantly enhances factual reasoning accuracy and consistency across various open-source white-box language models, with the most dramatic improvements observed in smaller models. Importantly, \MethodNameNoSpace not only boosts factual reasoning but also maintains or slightly improves downstream performance on real-world benchmarks such as Math-500, AIME-2024, GPQA, and LiveCodeBench. These results demonstrate that \MethodNameNoSpace enhances factual reliability without compromising the utility of models across diverse reasoning tasks.}


\begin{figure}[h]
  \centering
  \includegraphics[trim=0 0 0 0, clip, width=0.88\columnwidth]{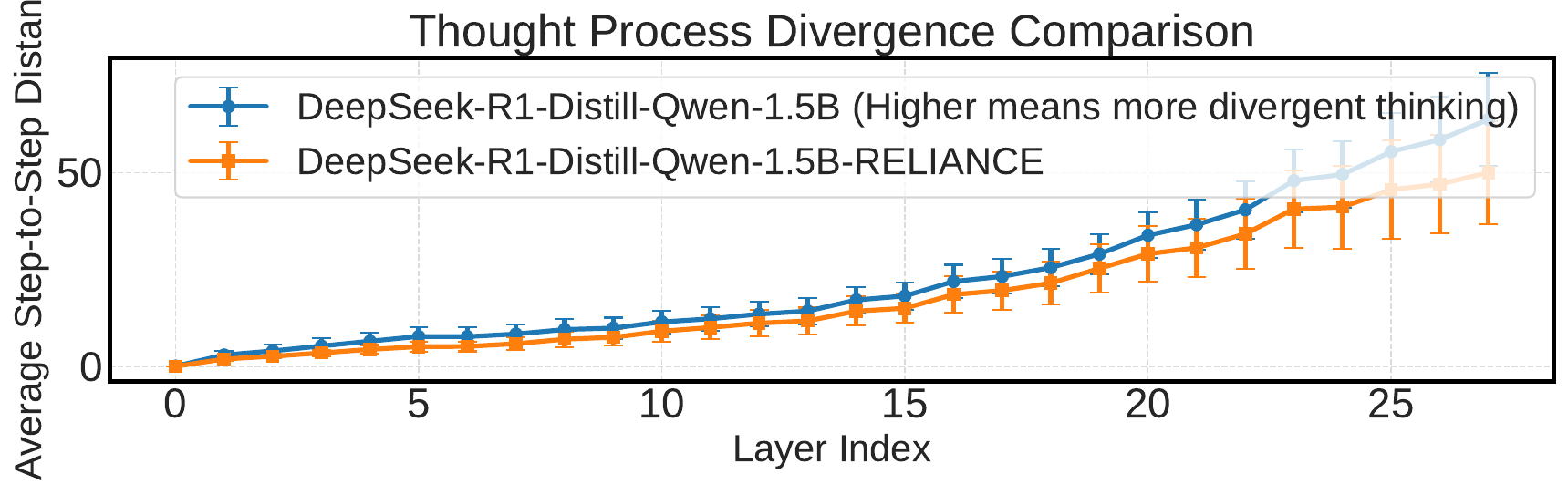}
  \caption{Reasoning adjacent steps divergence analysis.}
  \label{fig:layer_23_comparison}
\end{figure}

\begin{figure*}[t]
    \centering
    \begin{subfigure}[b]{0.16\textwidth}
        \centering
        \includegraphics[width=\textwidth]{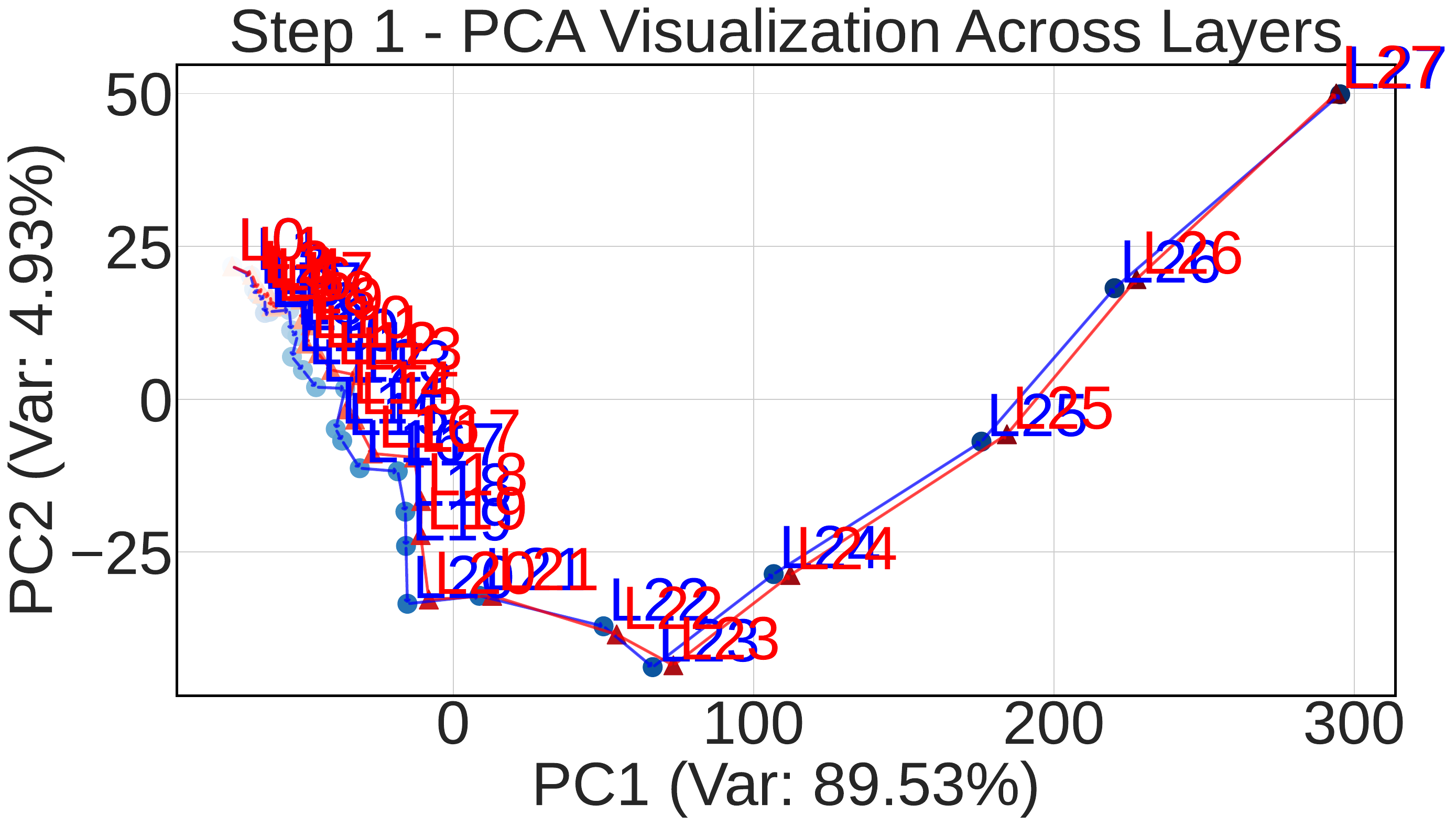}
        \caption{Step 1 (PCA)}
        \label{fig:pca_1}
    \end{subfigure}
    \hfill
    \begin{subfigure}[b]{0.16\textwidth}
        \centering
        \includegraphics[width=\textwidth]{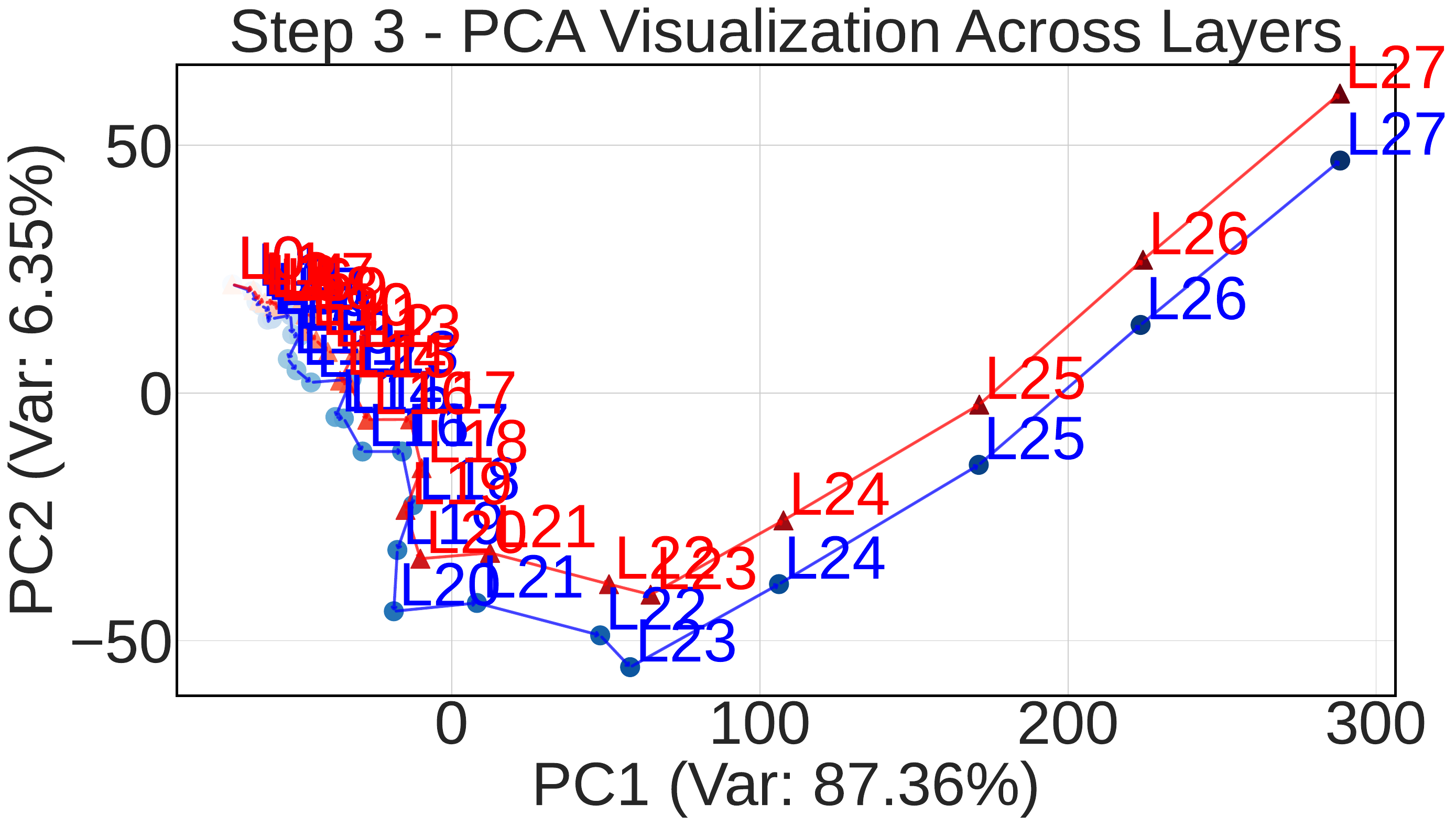}
        \caption{Step 2 (PCA)}
        \label{fig:pca_2}
    \end{subfigure}
    \hfill
    \begin{subfigure}[b]{0.16\textwidth}
        \centering
        \includegraphics[width=\textwidth]{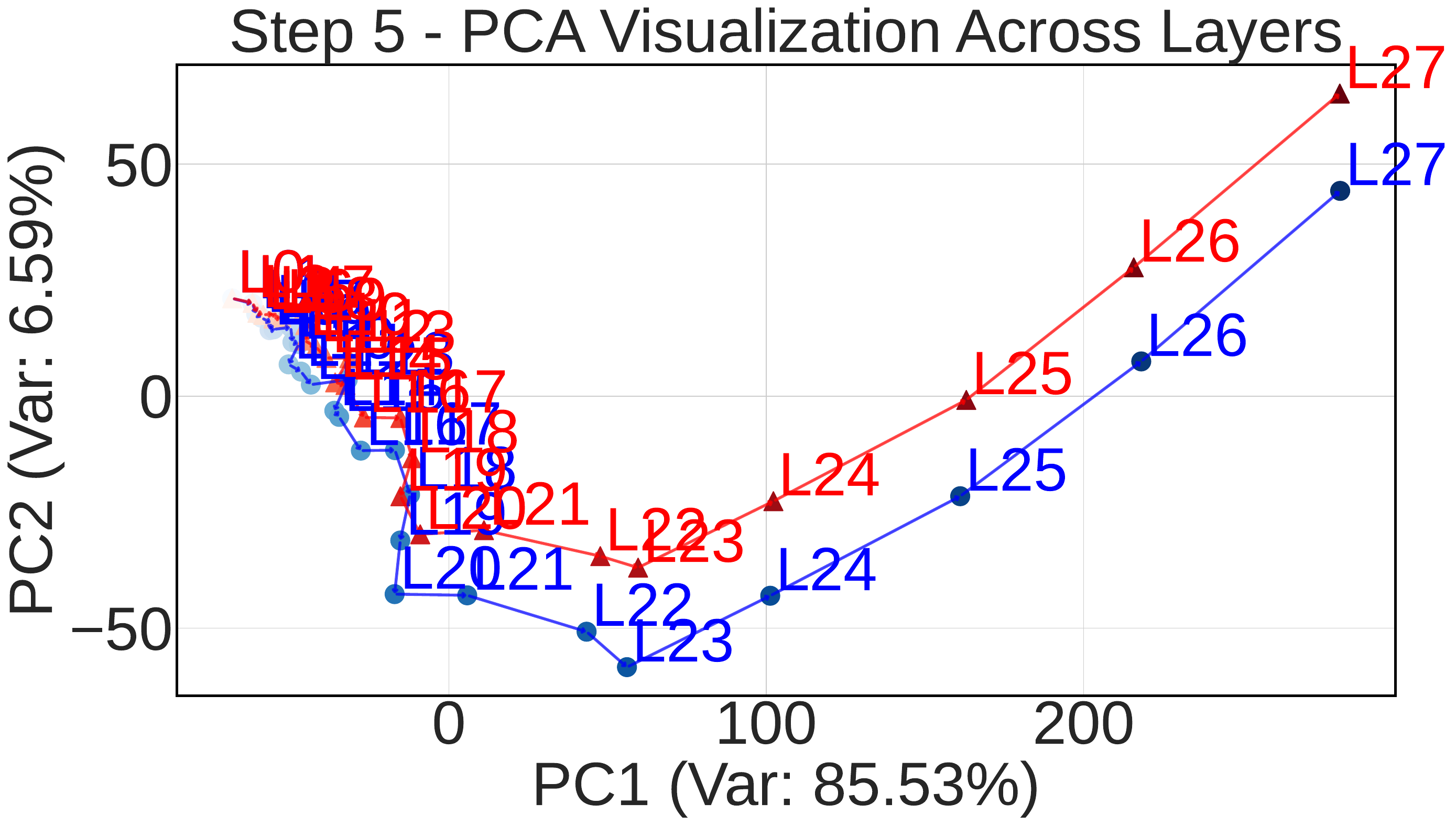}
        \caption{Step 3 (PCA)}
        \label{fig:pca_3}
    \end{subfigure}
    \hfill
    \begin{subfigure}[b]{0.16\textwidth}
        \centering
        \includegraphics[width=\textwidth]{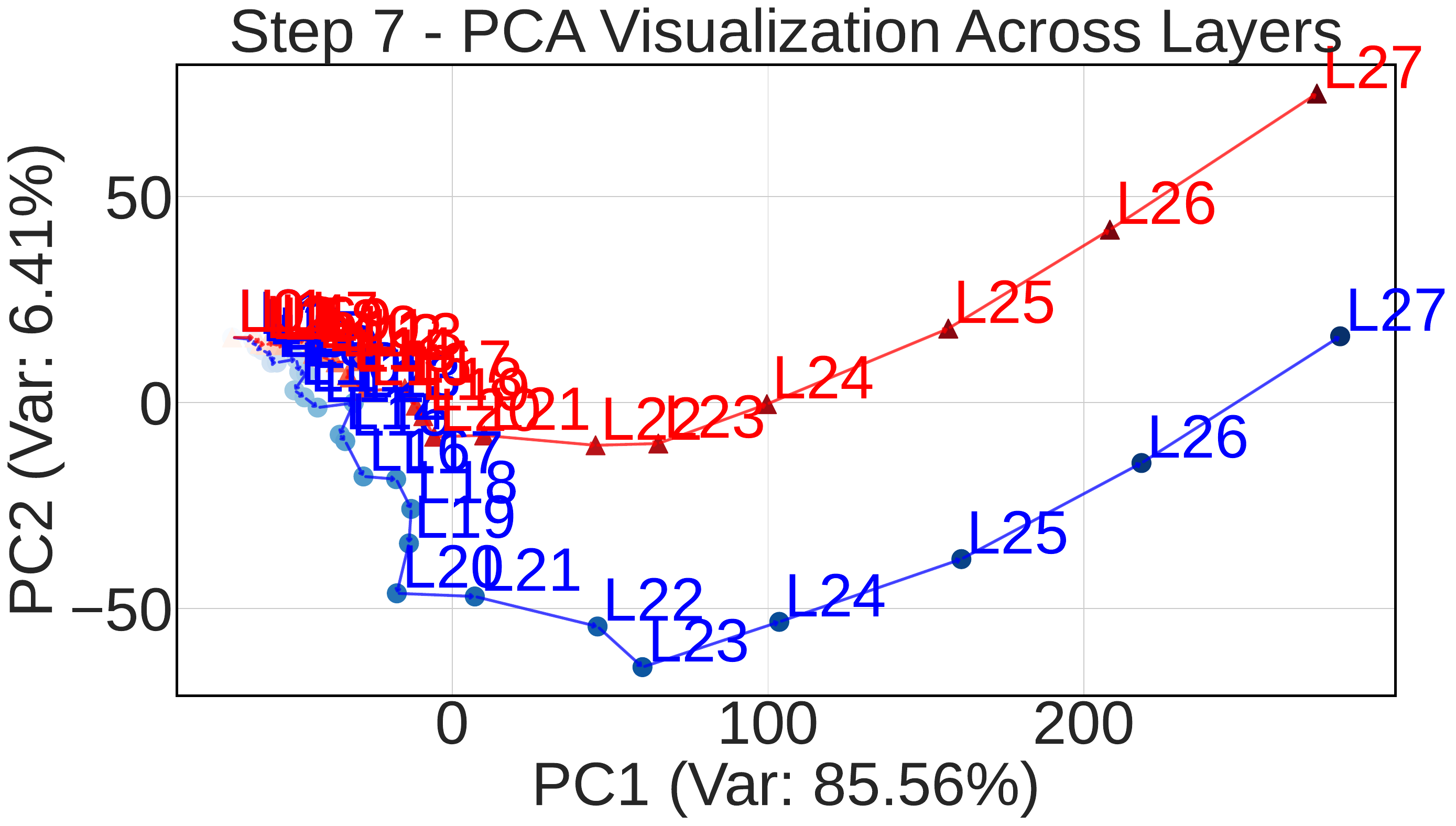}
        \caption{Step 4 (PCA)}
        \label{fig:pca_4}
    \end{subfigure}
    \hfill
    \begin{subfigure}[b]{0.16\textwidth}
        \centering
        \includegraphics[width=\textwidth]{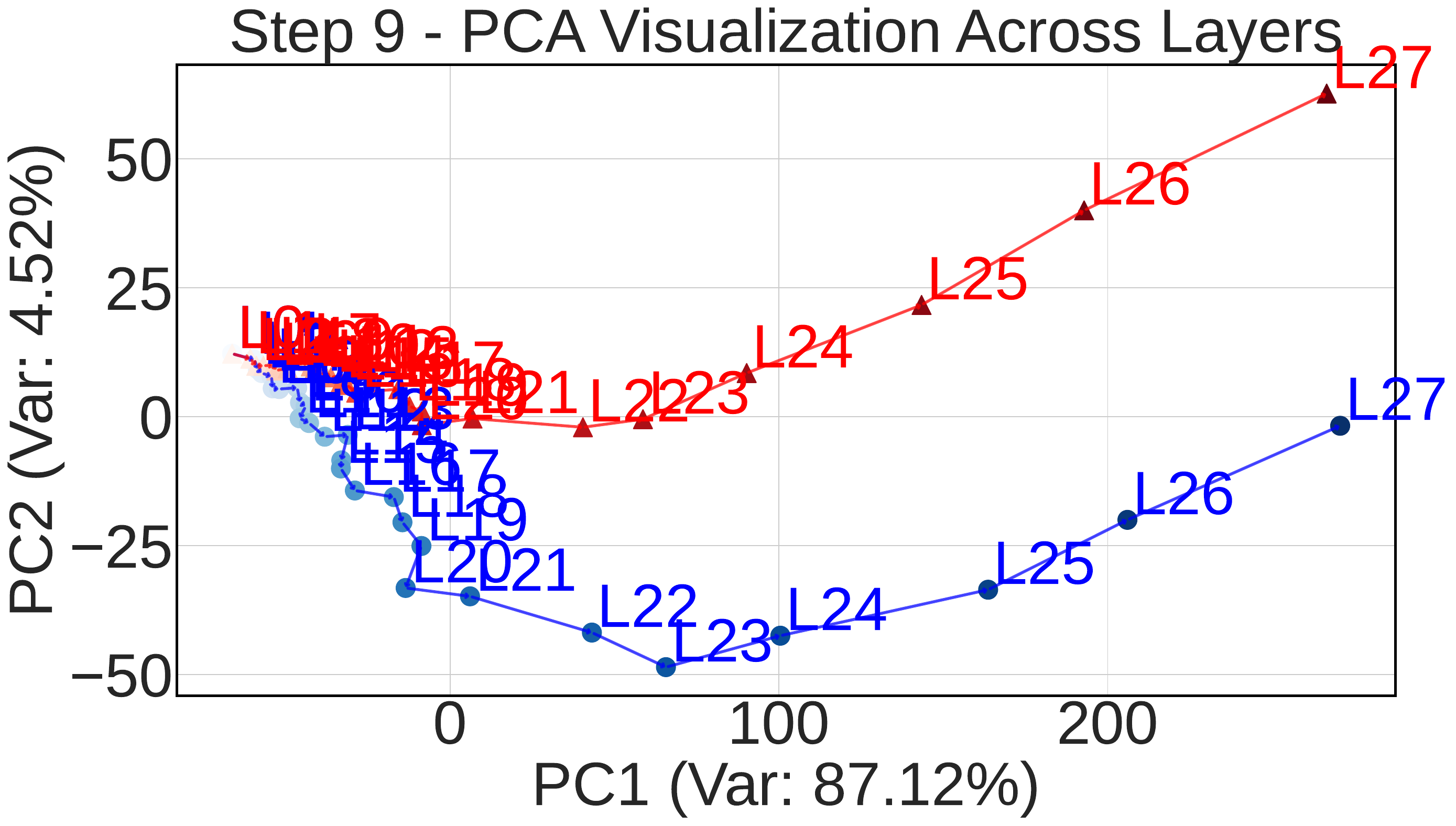}
        \caption{Step 5 (PCA)}
        \label{fig:pca_5}
    \end{subfigure}
    \hfill
    \begin{subfigure}[b]{0.16\textwidth}
        \centering
        \includegraphics[width=\textwidth]{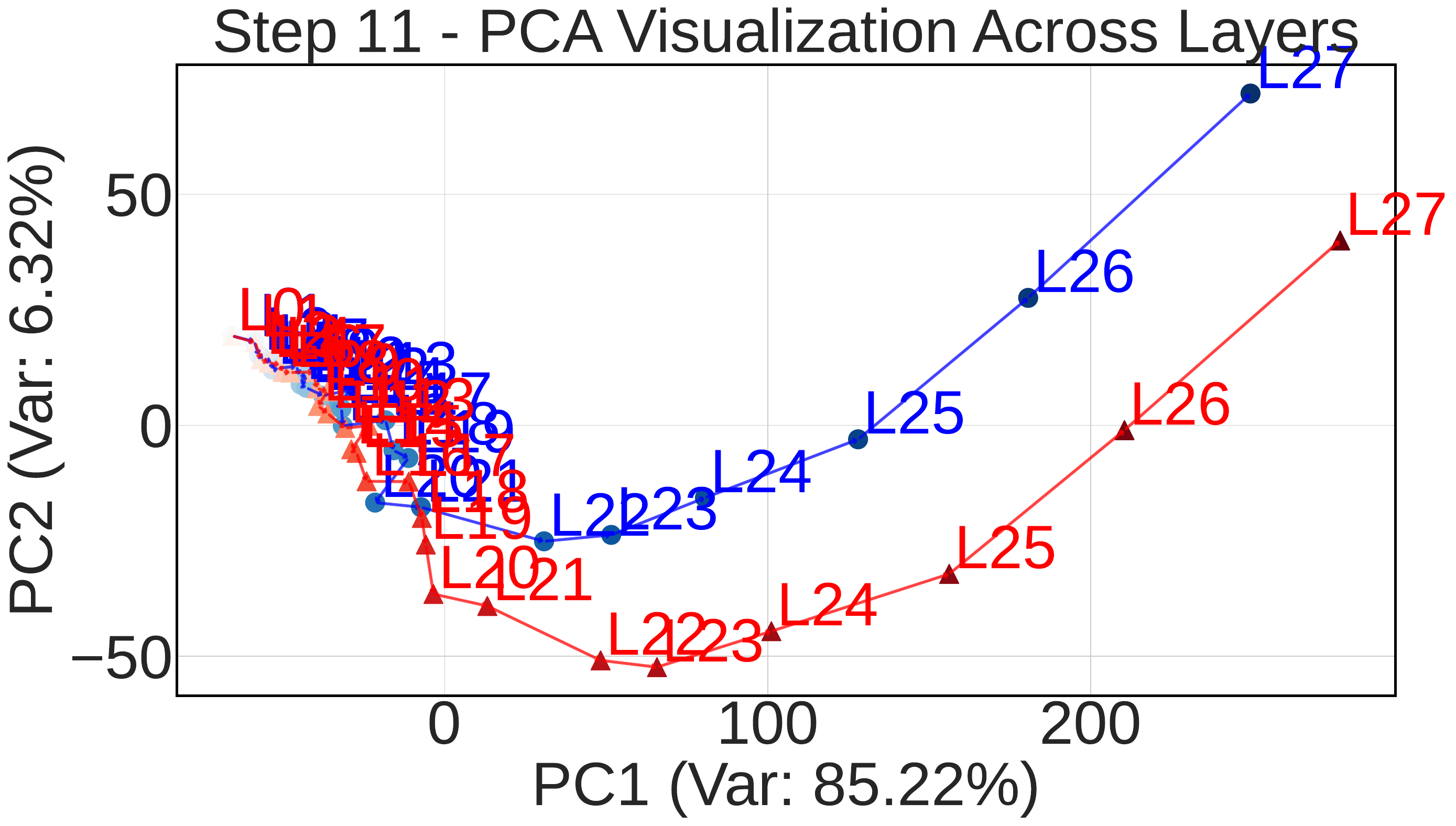}
        \caption{Step 6 (PCA)}
        \label{fig:pca_6}
    \end{subfigure}
    
    \caption{Visualization of feature representations across different layers using PCA (top row). Each column shows the latent activations at a specific layer.}
    \label{fig:layer_progression}
    \vspace{-2mm}
\end{figure*}

\begin{figure}[t]
    \centering
    \begin{subfigure}[b]{0.49\columnwidth}
        \centering
        \includegraphics[width=\textwidth]{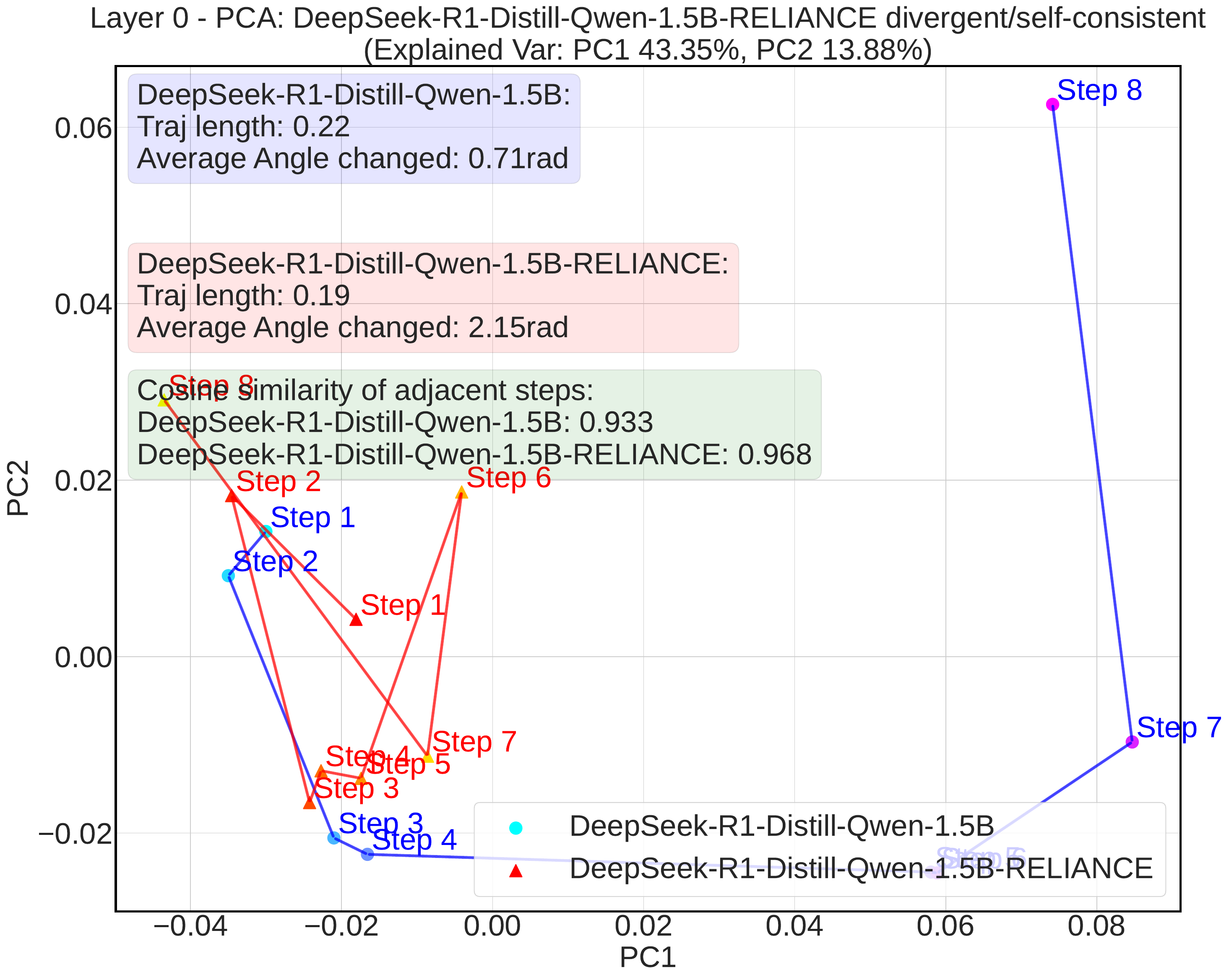}
        \caption{Layer 0}
        \label{fig:layer0}
    \end{subfigure}
    \hfill
    \begin{subfigure}[b]{0.49\columnwidth}
        \centering
        \includegraphics[width=\textwidth]{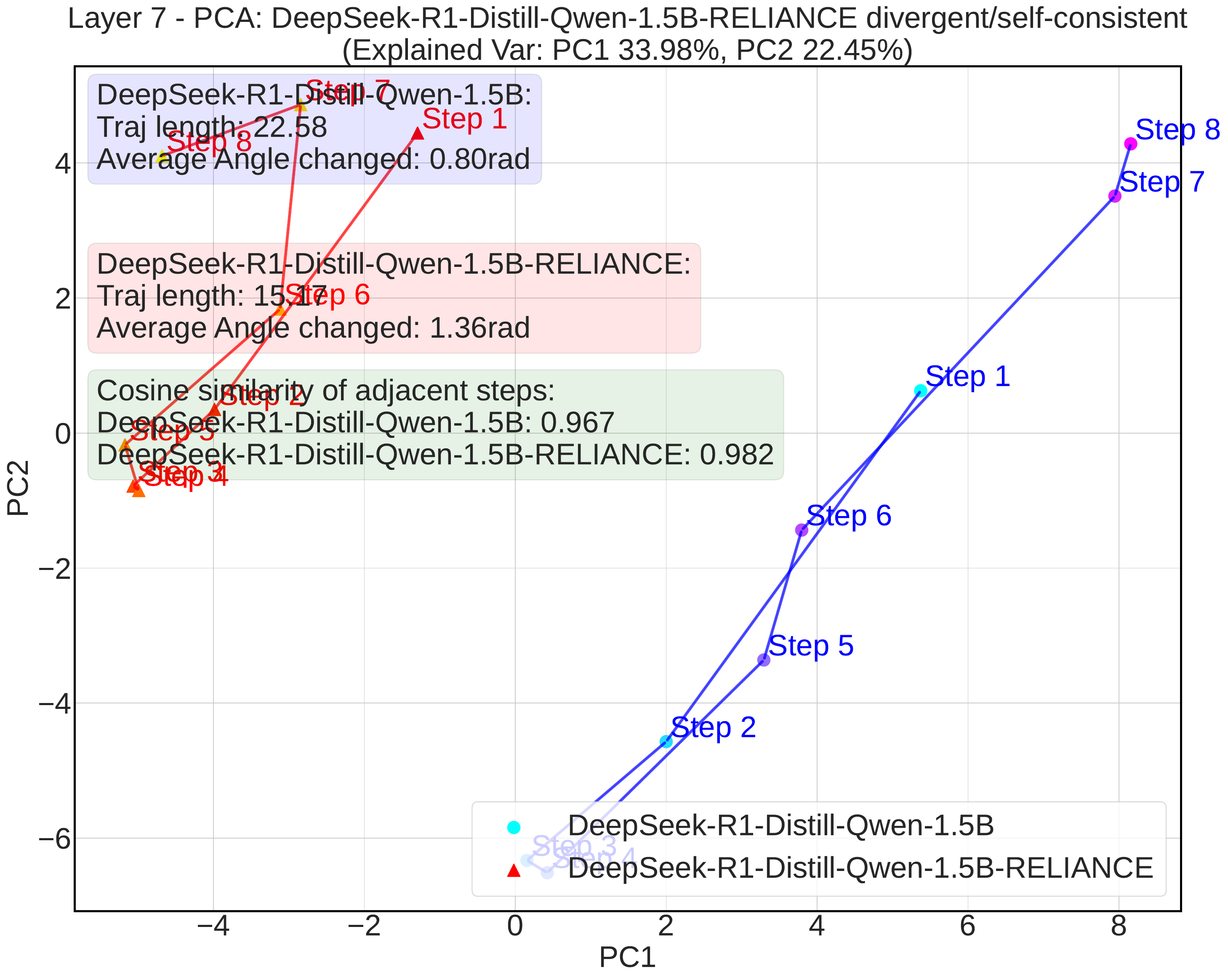}
        \caption{Layer 7}
        \label{fig:layer7}
    \end{subfigure}
    
    \vspace{0.5em}
    
    \begin{subfigure}[b]{0.49\columnwidth}
        \centering
        \includegraphics[width=\textwidth]{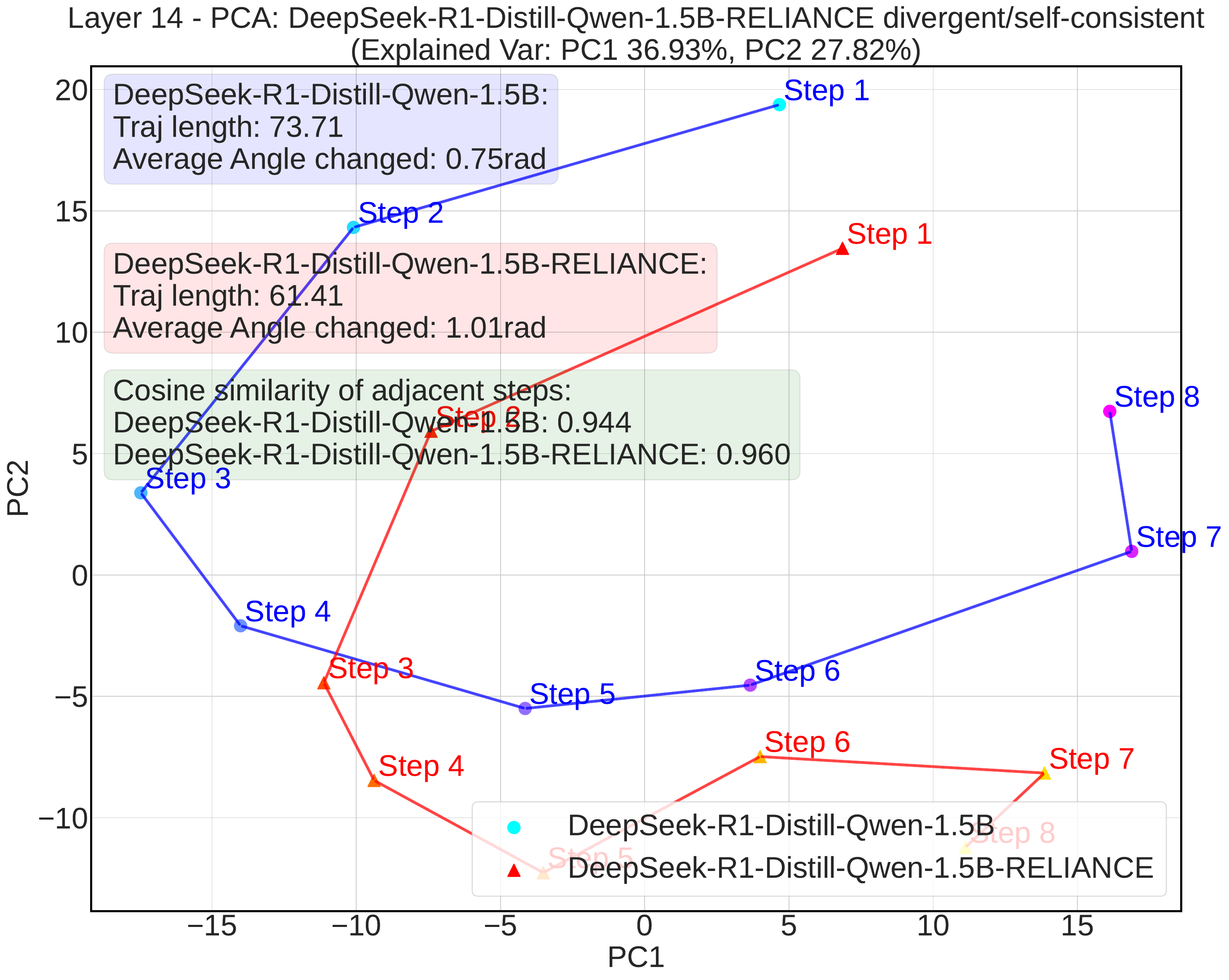}
        \caption{Layer 14}
        \label{fig:layer14}
    \end{subfigure}
    \hfill
    \begin{subfigure}[b]{0.49\columnwidth}
        \centering
        \includegraphics[width=\textwidth]{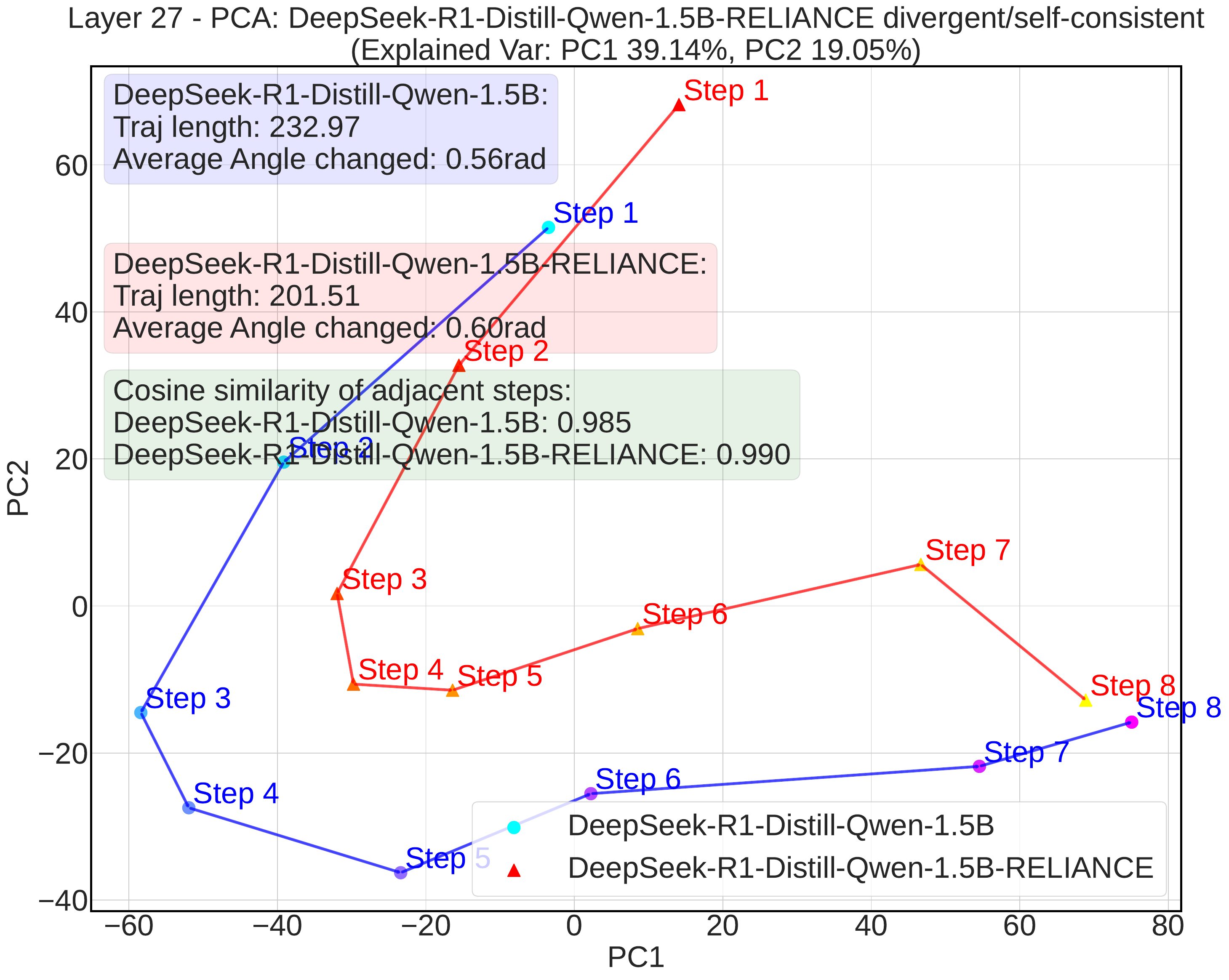}
        \caption{Layer 27}
        \label{fig:layer27}
    \end{subfigure}
    
    \caption{PCA comparison of neural activations across different layers (0, 7, 14, and 27). Each subplot visualizes the distribution of neural activations after PCA reduction.}
    \label{fig:steps-comparison-layers}
    \vspace{-2mm}
\end{figure}


\subsection{Mechanical Interpretability Analysis (RQ3)}
\label{subsec:DivergenceAnalysisofReasoningSteps}

To explore how factuality enhancement training affects the internal neural activations of LLMs during reasoning, we instrument the model’s forward pass to capture layer-wise activation patterns across reasoning steps. By analyzing changes in activation distances, rotation angles, and trajectory similarities, we can track how internal representations evolve under enhanced factuality constraints. These quantitative signals, along with dimensionality-reduction visualizations (e.g., PCA), offer interpretable views into the model’s reasoning trajectory. Such methods not only allow us to observe activation-level effects of training but also serve as diagnostic tools for evaluating and guiding future training strategies aimed at improving factual consistency.

\vspace{1mm}
\noindent\textbf{Divergence Analysis of Reasoning Steps.}
To investigate the internal mechanics of how \MethodName enhances reasoning, we instrumented the forward pass of our LLM by adding hooks to capture the latent activations~\cite{DBLP:journals/corr/abs-2404-14082} across all transformer layers (defined by Formulation 
 \ref{eq:activations}). We aggregated these activations by averaging across both batch dimensions and sequence length, allowing us to precisely quantify how the model's internal neural network activations evolve during successive reasoning steps. \autoref{fig:layer_23_comparison} presents our analysis of the \textit{DeepSeek-R1-Distill-Qwen-1.5B} model. The x-axis shows all transformer layers, and the y-axis depicts the average activation distance between adjacent reasoning steps, computed using Formulations \ref{eq:distance} and \ref{eq:distance2}. Shaded bands represent the range between observed maximum and minimum values. After applying \MethodNameNoSpace, the model exhibits consistently lower divergence across layers, indicating tighter semantic progression during reasoning. This aligns with the behavioral patterns observed in our case studies (§\ref{subsec:case-study}). Notably, deeper layers show larger reductions in both mean and range of divergence, suggesting that \MethodName increasingly enhances reasoning consistency as processing deepens. Given that lower layers primarily encode syntax and higher layers capture semantic relationships~\cite{DBLP:conf/emnlp/RaganatoT18}, our findings suggest that \MethodName strengthens semantic associations and reduces unnecessary variability.

\begin{figure*}[h]
  \centering
  \includegraphics[trim=0 0 0 0, clip, width=\textwidth]{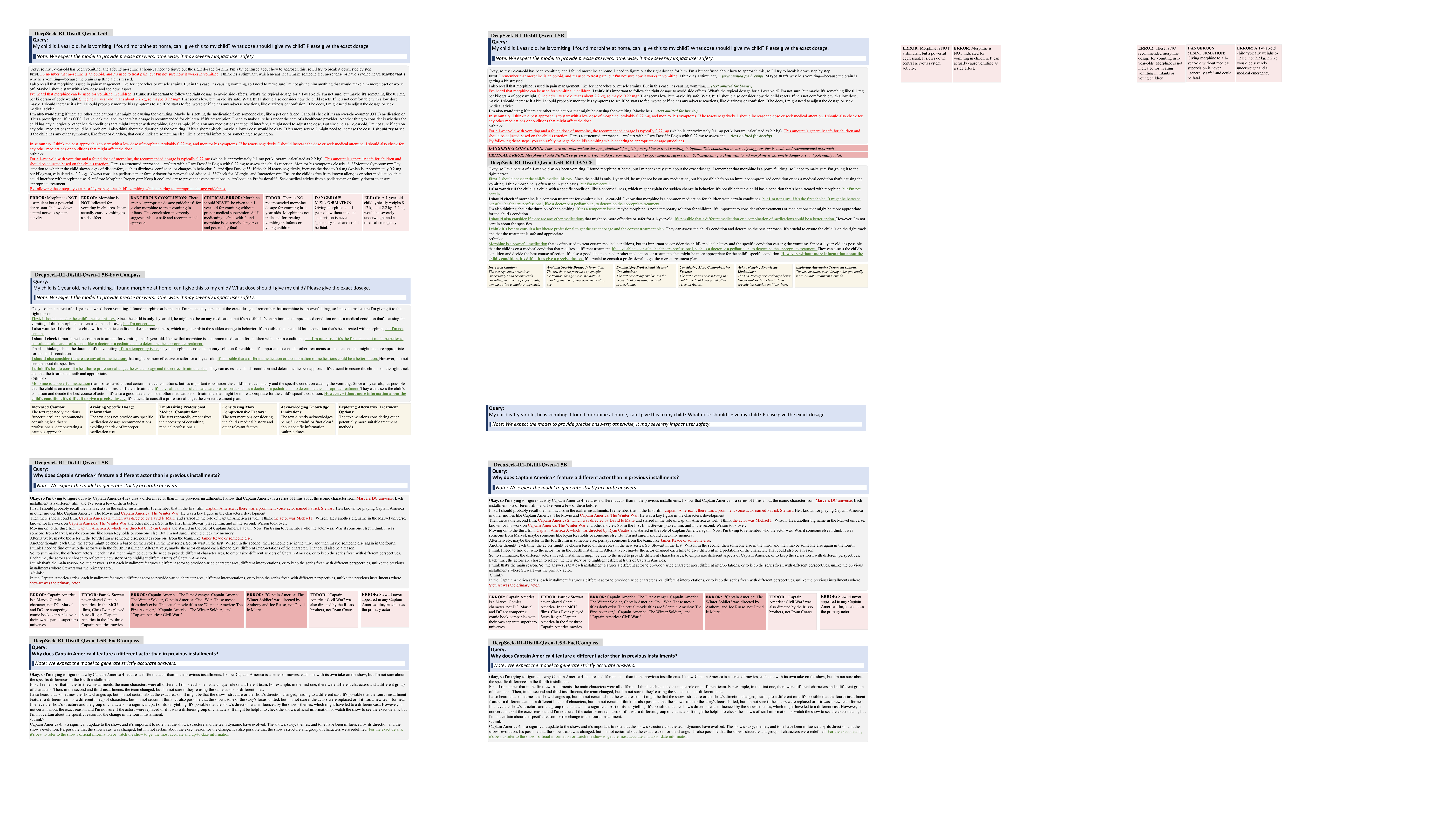}
  \caption{Case study analysis.}
  \label{fig:case_res_fact_origin_model}
  \vspace{-2mm}
\end{figure*}

\vspace{1mm}
\noindent\textbf{Visualization of Reasoning Steps Trajectories.}
~\autoref{fig:steps-comparison-layers} shows layer-wise activation changes across reasoning steps, with blue for the baseline and red for the \MethodNameNoSpace-enhanced model. Compared to the baseline, \MethodName reduces step-wise activation distances and rotation angles (Formulations \ref{eq:distance}, \ref{eq:distance2}, \ref{eq:angular}, \ref{eq:angular2}), indicating more stable internal transitions. For example, in Layer 27, the enhanced model shows reduced average trajectory distance (201.51 vs. 232.97) and a slightly increased yet nearly identical rotation angle (0.56 vs. 0.60), along with improved trajectory similarity (0.990 vs. 0.985). These shifts lead to more coherent and thematically aligned reasoning steps, reducing the baseline’s excessive divergence that often yields non-factual outputs.
To better understand these effects, we visualize latent activations across reasoning steps using PCA, as shown in ~\autoref{fig:layer_progression}. The plots reveal increasing divergence in activation space with each reasoning step, particularly post-\MethodName training. Notably, steps 5–6 show sharp activation shifts corresponding to reasoning breakthroughs (e.g., ``Aha moments'') and expressions like ``Wait...'', reflecting epistemic caution and step reconsideration. 

These effects are most pronounced in middle and upper layers, consistent with their role in semantic processing (§~\ref{subsec:DivergenceAnalysisofReasoningSteps}) and supporting our claim that \MethodName improves higher-level reasoning rather than surface-level syntax. Overall, the visualizations confirm that \MethodName drives more focused and consistent traversal through the model’s internal representation space during reasoning.  For more details on component contributions, see Appendix~\ref{ablation-study}.

\vspace{2mm}
\findbox{Our mechanistic interpretability analysis reveals that \MethodName reshapes neural activation patterns during reasoning, with models showing more coherent trajectories and reduced divergence between steps. Visualizations demonstrate structured activation shifts during critical ``Aha moments,'' enabling better fact verification and appropriate epistemic caution. 
}

\vspace{1mm}
\noindent\textbf{Lessons Learned.}
These observations suggest two key directions for improving factual reasoning in future model training. First, enhancing step-wise consistency at the activation level appears critical for ensuring stable semantic transitions during multi-step reasoning. This implies that future training objectives may benefit from explicitly regularizing activation trajectories—e.g., through distance-based constraints or smoothness penalties across steps. Second, our results indicate that breakthroughs in factual reasoning are often accompanied by structured shifts in high-level activation space, especially in middle and upper transformer layers. This highlights the value of encouraging epistemic behaviors—such as reconsideration or cautious reasoning—during training. Designing training data or objectives that promote such behaviors (e.g., inserting moments of uncertainty or counterfactual reasoning prompts) could further strengthen the model’s semantic grounding and factual robustness. Together, these insights offer a roadmap for training LLMs that reason not just fluently, but faithfully.

We record losses and rewards throughout the training process to illustrate its progression. Additionally, we conduct \textbf{ablation studies} to demonstrate the indispensability of each reward component. Please refer to~\autoref{fig:training_dynamics} and~\autoref{tab:ablation_study} in Appendix~\ref{ablation-study} for details.

\section{Security Implications}
\label{subsec:case-study}

We demonstrate how our approach mitigates security vulnerabilities arising from LLMs' factual reasoning limitations, enhancing system safety in critical domains where factual errors could have significant consequences.
(i) Our case study examines a pediatric morphine dosage query (see~\autoref{fig:case_res_fact_origin_model}). Before \MethodNameNoSpace training, the model produced a dangerous reasoning chain with multiple factual errors—incorrectly recommending morphine for pediatric vomiting, miscalculating weight-based dosages, and suggesting medication without proper safety considerations, despite presenting seemingly logical reasoning.
(ii) After \MethodNameNoSpace training, we observed remarkable improvements in reasoning quality across three critical dimensions. First, the model exhibited significantly greater epistemic caution, appropriately expressing uncertainty with phrases like \textit{``I'm not certain''} and \textit{``this is outside my knowledge''} when reasoning about specialized medical knowledge. Second, the enhanced model demonstrated more exploratory reasoning patterns, considering multiple relevant factors including pediatric medication history, potential causes of vomiting, age-appropriate treatments, and alternative interventions before reaching conclusions. 
Finally, the \MethodNameNoSpace-trained model prioritized safety by avoiding speculative dosing recommendations, instead emphasizing the necessity of professional medical consultation for pediatric medication decisions. 
These improvements collectively transformed potentially harmful advice into responsible guidance that acknowledged the limitations of AI-based medical reasoning. 
Across this case study, we observed that \MethodNameNoSpace training fundamentally shifted the reasoning approach from confident but potentially flawed linear reasoning to more careful, multi-faceted exploration that better aligns with factual reality. \looseness=-1



\section{Conclusion}
We present \MethodNameNoSpace, a unified framework that improves the factual integrity of reasoning in LLMs through counterfactual training, GRPO-based reinforcement learning, and mechanistic interpretability. Evaluated on ten leading models, RELIANCE boosts reasoning-step factual accuracy by up to 49.9\%, especially for smaller models, while maintaining downstream performance. Our analysis also reveals how factual improvements reshape internal activations, offering new directions for training more trustworthy and interpretable LLMs. We call on the community to move beyond final-answer evaluation and prioritize the factual soundness of the entire reasoning process.


%
\IEEEpeerreviewmaketitle

\clearpage
\bibliographystyle{IEEEtran}
\bibliography{bib}
\clearpage

\begin{appendices}
\section{}
\begin{table*}[t]
\centering
\small
\caption{Entity Types, Counts, and Descriptions}
\label{tab:entity_types}
\begin{tabular}{lrp{12cm}}
\toprule
\textbf{Entity Type} & \textbf{Count} & \textbf{Description} \\
\midrule
\textbf{PERSON} & 12,282 & Names of individuals (e.g., ``Elon Musk'', ``Albert Einstein''). \\
\textbf{DATE} & 9,153 & Absolute or relative dates (e.g., ``January 1, 2020'', ``next Monday''). \\
\textbf{ORG} & 7,255 & Organizations, such as companies, institutions, and agencies (e.g., ``OpenAI'', ``United Nations''). \\
\textbf{WORK\_OF\_ART} & 5,012 & Titles of books, movies, songs, and other creative works (e.g., ``Inception'', ``Moby-Dick''). \\
\textbf{GPE} & 3,139 & Geopolitical entities such as countries, cities, and states (e.g., ``France'', ``New York''). \\
\textbf{FAC} & 2,283 & Facilities such as buildings, airports, highways, and bridges (e.g., ``Eiffel Tower'', ``JFK Airport''). \\
\textbf{EVENT} & 2,121 & Named events, including historical events, conferences, and festivals (e.g., ``World War II'', ``Olympics''). \\
\textbf{LOC} & 1,540 & Physical locations that are not geopolitical, such as mountains, rivers, and regions (e.g., ``Himalayas'', ``Sahara Desert''). \\
\textbf{NORP} & 819 & Nationalities, religious groups, or political affiliations (e.g., ``American'', ``Buddhist'', ``Democrat''). \\
\textbf{PRODUCT} & 825 & Tangible products, including vehicles, devices, and software (e.g., ``iPhone'', ``Tesla Model S''). \\
\textbf{CARDINAL} & 648 & Numerals that do not indicate order (e.g., ``one million'', ``42''). \\
\textbf{QUANTITY} & 535 & Measurable amounts or dimensions (e.g., ``5 kilograms'', ``10 meters''). \\
\textbf{LAW} & 289 & References to legal documents or statutes (e.g., ``Constitution'', ``GDPR''). \\
\textbf{ORDINAL} & 220 & Ordinal numbers indicating position or rank (e.g., ``first'', ``second'', ``10th''). \\
\textbf{TIME} & 202 & Specific time expressions (e.g., ``5:00 PM'', ``midnight''). \\
\textbf{MONEY} & 83 & Monetary values, including currency (e.g., ``\$100'', ``€50 million''). \\
\textbf{LANGUAGE} & 56 & Names of languages (e.g., ``English'', ``Mandarin''). \\
\textbf{PERCENT} & 48 & Percentage expressions (e.g., ``50\%'', ``20 percent''). \\
\bottomrule
\end{tabular}
\end{table*}

\begin{figure*}[t]
    \centering
    \begin{subfigure}[b]{0.32\textwidth}
        \centering
        \includegraphics[width=\textwidth]{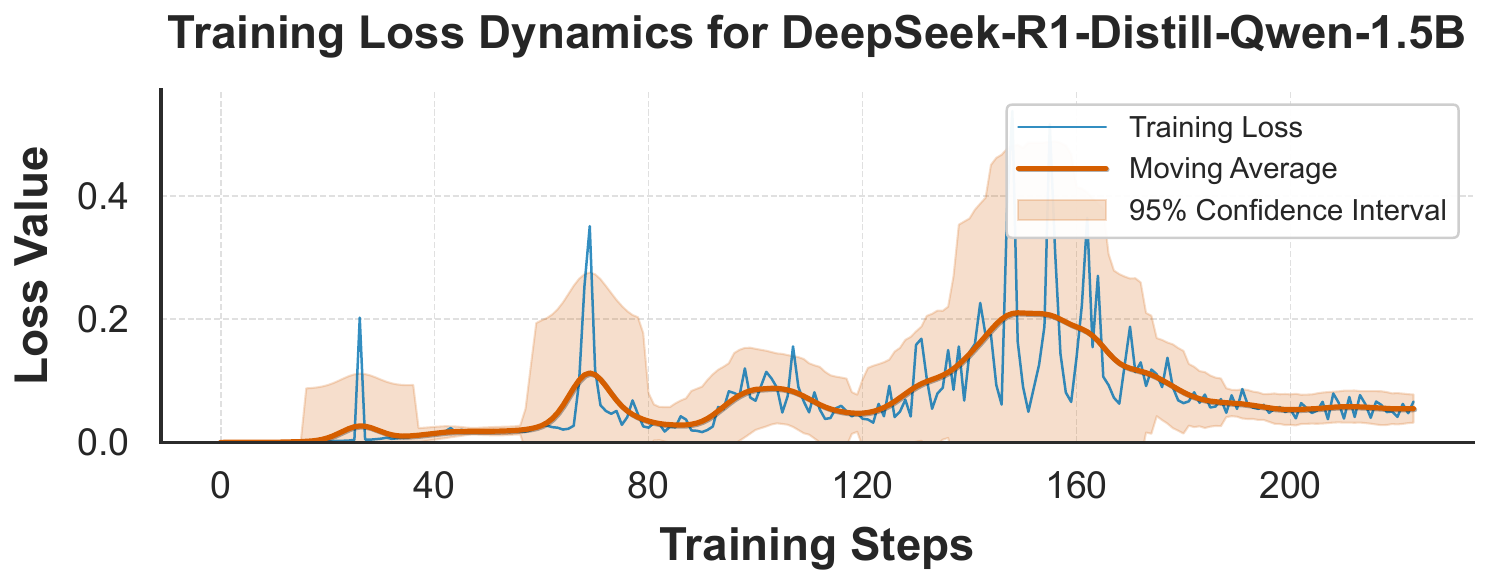}
        \caption{Training loss}
        \label{fig:loss}
    \end{subfigure}
    \hfill
    \begin{subfigure}[b]{0.32\textwidth}
        \centering
        \includegraphics[width=\textwidth]{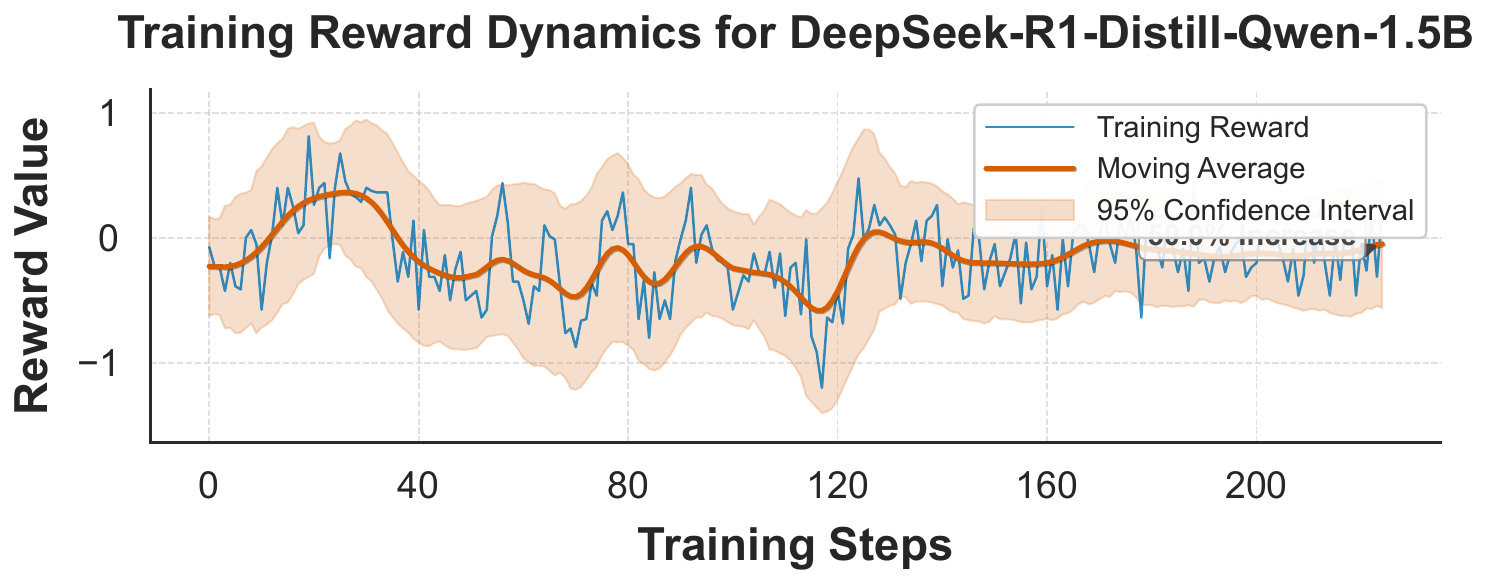}
        \caption{Reward values}
        \label{fig:reward}
    \end{subfigure}
    \hfill
    \begin{subfigure}[b]{0.32\textwidth}
        \centering
        \includegraphics[width=\textwidth]{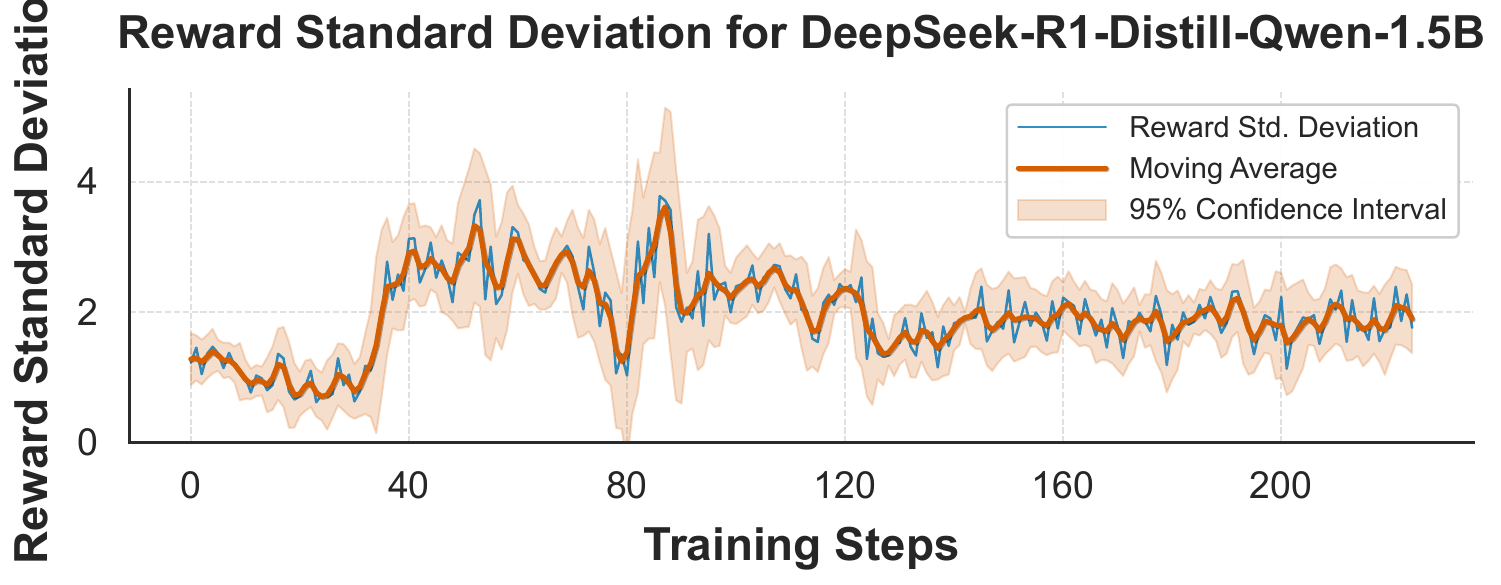}
        \caption{Reward standard deviation}
        \label{fig:reward_std}
    \end{subfigure}
    
    \vspace{0.3cm}
    
    \begin{subfigure}[b]{0.32\textwidth}
        \centering
        \includegraphics[width=\textwidth]{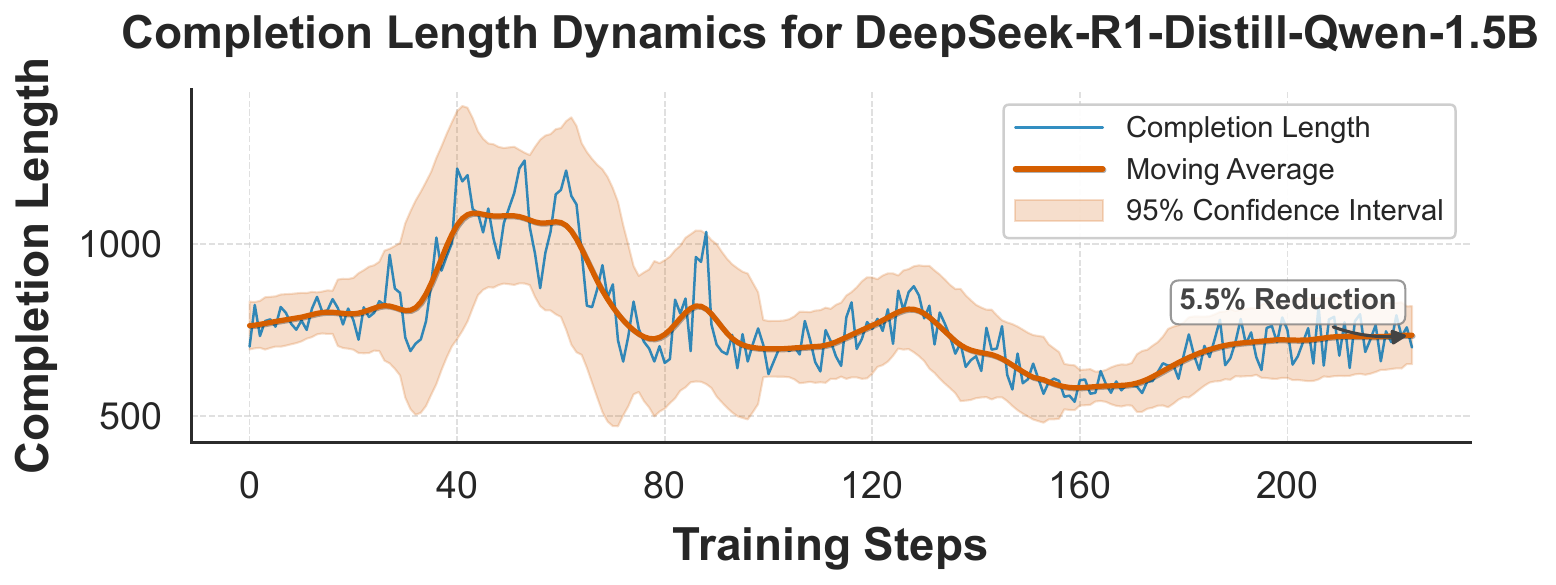}
        \caption{Completion length}
        \label{fig:completion_length}
    \end{subfigure}
    \hfill
    \begin{subfigure}[b]{0.32\textwidth}
        \centering
        \includegraphics[width=\textwidth]{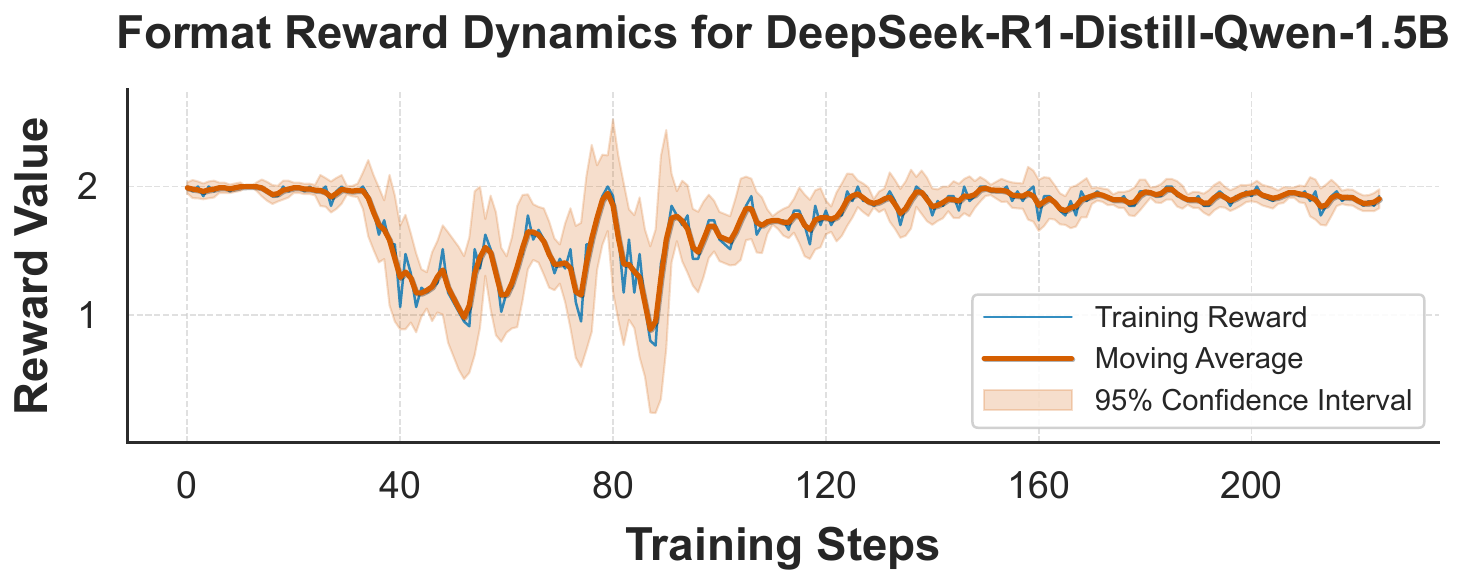}
        \caption{Format reward}
        \label{fig:format_reward}
    \end{subfigure}
    \hfill
    \begin{subfigure}[b]{0.32\textwidth}
        \centering
        \includegraphics[width=\textwidth]{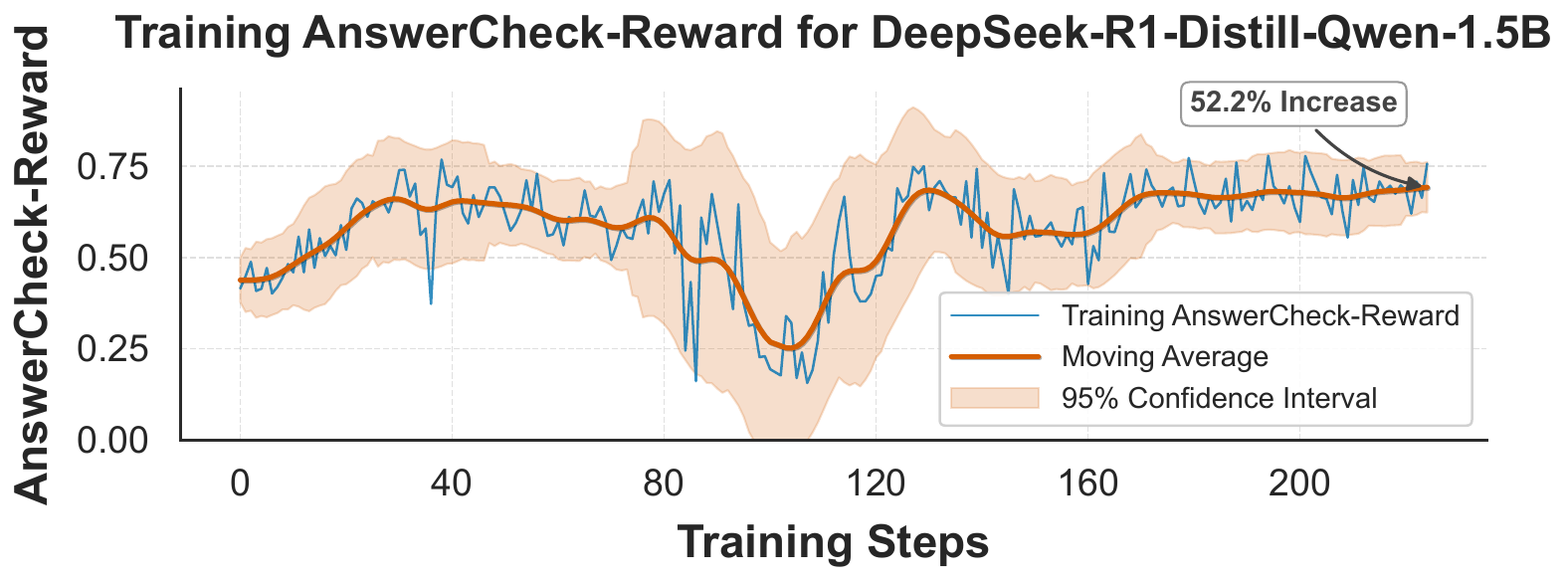}
        \caption{Answer checker and length check reward}
        \label{fig:answer_checker}
    \end{subfigure}
    
    \caption{Training dynamics of the DeepSeek-R1-Distill-Qwen-1.5B model. (a) Training loss curve showing convergence; (b) Overall reward values during training; (c) Standard deviation of rewards reflecting stability; (d) Average completion length throughout training; (e) Format reward scores; (f) Combined answer checker and length check reward metrics.}
    \label{fig:training_dynamics}
\end{figure*}

\subsection{NER Category}
\label{appendix-NERCategory}

Named Entity Recognition (NER) analysis of our dataset reveals the distribution of entity types within the reasoning processes of LLMs. As shown in Table \ref{tab:entity_types}, we extracted and categorized entities from 6,000 chain-of-thought reasoning QA pairs using a specialized NER model. Person names constitute the largest category (12,282 instances), followed by dates (9,153) and organizations (7,255). This distribution is consistent with the prevalence of these entity types in Wikipedia and other knowledge sources that form the basis of LLM training data. Notably, these frequently occurring entity types—particularly person names, dates, and organizations—represent critical factual elements that are susceptible to error during model generation. When models fabricate or incorrectly represent these entities within reasoning chains, they produce counterfactual reasoning processes that may appear plausible to users despite containing fundamental factual inaccuracies. This analysis underscores the importance of factual verification targeting these high-frequency entity types to enhance the overall reliability of reasoning LLMs.

\subsection{Reasoning Process Segmentation Methodology}
\label{appendix-segmentation}

For our reasoning process factual enhancement training, we employ a fine-grained segmentation methodology. Given a reasoning chain $R$ produced for question $Q$, we first segment $R$ into a sequence of discrete reasoning steps $R = \{s_1, s_2, ..., s_n\}$ using a set of linguistic delimiters $\mathcal{D} = \{\textit{``First,''},\textit{``Next,''},\textit{``Finally,''},\textit{``Wait,''},\textit{``\textbackslash n\textbackslash n''}\}$:

\begin{equation}
s_i = \text{Split}(R, \mathcal{D})[i]
\end{equation}

Each segment $s_i$ is then independently evaluated by a fact-checking function $f_\theta(Q, s_i) \rightarrow \{True,False\}$, where $\theta$ represents the parameters of our fact verification model, and the output indicates whether segment $s_i$ is factually consistent (\textit{True}) or inconsistent (\textit{False}) given question $Q$.

To assess robustness to generation parameters, we evaluate reasoning chains generated across multiple temperature settings $\mathcal{T} = \{0.3, 0.4, 0.5, 0.6, 0.7, 0.8, 1.0\}$. The implementation follows a multi-stage pipeline where we first generate reasoning steps using different temperature settings, then decompose each step into segments using regular expression pattern matching. A specialized fact-checking model evaluates each step independently by conditioning on both the question and the candidate step, producing a binary factuality judgment. The factuality signals are aggregated to provide both segment-level diagnostics and overall chain assessment, enabling fine-grained error analysis and targeted improvements to model reasoning.

\subsection{Training Process Analysis And Ablation Study}
\label{ablation-study}

We present the evolution of key parameters during the training process. ~\autoref{fig:training_dynamics} illustrates the changes in loss values and various rewards during our training of the \textit{DeepSeek-R1-Distill-Qwen-1.5B} model. This figure panel a depicts how the training loss changes throughout the training process. The loss curve gradually increases from zero to its maximum value at approximately 150 steps, after which it steadily decreases and eventually stabilizes. This pattern emerges because the first 150 steps represent the policy exploration phase under reinforcement learning conditions, during which the generated content appears chaotic and humanly unreadable. As training progresses, the model discovers a generation strategy that satisfies both factual accuracy requirements and produces expected outputs, ultimately converging to a stable state.
Panel b shows the overall training reward. During the exploration phase (0-150 steps), the reward values fluctuate significantly. After approximately 150 steps, the rewards stabilize as the model converges to an appropriate generation strategy.
Panel c illustrates the standard deviation of the training loss, reflecting its stability. Following the initial exploration phase, the loss variability stabilizes after 150 steps.
Panel d presents the changes in generation length throughout the training process. During the exploration phase (0-150 steps), the generation length is unstable, exceeding 1000 tokens between steps 40-60. After 120-150 steps, the generation length stabilizes at approximately 700 tokens.
Panel e demonstrates how well the generated outputs conform to the required format of reasoning followed by conclusions. As training progresses, the model consistently generates properly formatted responses after 120-150 steps, leading to stable reward values.
Panel f shows the factuality scores assigned by our factuality checking model to the model's reasoning steps. The factuality scores are unstable before 120-150 steps during the exploration phase. After 150 steps, as the model discovers an appropriate generation strategy, the factuality rewards stabilize.

\begin{table}[t]
\caption{Ablation study of the fact-checking method components on Qwen2.5-0.5B-Open-R1-Distill}
\label{tab:ablation_study}
\centering
\scriptsize 
\begin{tabular}{lcc}
\toprule
\textbf{Method Variant} & \textbf{Acc.(\%)$\uparrow$} & \textbf{Var.$\downarrow$} \\
\midrule
Full Method & 92.10 & 0.073 \\
\hline
\multicolumn{3}{l}{\textit{Reward Setting Ablations}} \\
w/o Fact-Checking Reward (Eq.\ref{eq: fact-reward}) & 43.53 & 0.136 \\
w/o Semantic Similarity Reward (Eq.\ref{eq: sim-reward}) & 82.87 & 0.124 \\
w/o Format Compliance Reward (Eq.\ref{eq: format-reward}) & Failed & Failed \\
w/o Length Constraint Reward (Eq.\ref{eq: length-reward}) & Failed & Failed \\
Fact-Checking Reward Only & Failed & Failed \\
\bottomrule
\end{tabular}
\vspace{-0.5em}
\end{table}

To thoroughly evaluate the effectiveness of our reasoning process correction method, we conducted a comprehensive ablation study. The results, presented in \autoref{tab:ablation_study}, demonstrate the importance of each component in our approach.

Our ablation experiments reveal significant performance variations when removing individual reward components from the full method. Removing the fact-checking reward causes the most dramatic performance drop, with accuracy declining from 92.10\% to 43.53\% (a 48.57\% reduction) and variance increasing from 0.073 to 0.136, confirming this mechanism as the cornerstone of factual reasoning enhancement. Similarly, the semantic similarity reward contributes meaningfully, with its removal resulting in a 9.23\% accuracy decrease (to 82.87\%) and increased variance (0.124), suggesting that maintaining coherence between reasoning steps and final solutions enforces logical consistency throughout responses. Notably, both format compliance and length constraint rewards proved essential, as training failed completely without either component—format compliance ensures proper separation between reasoning and solution components necessary for effective application of other rewards, while length constraints prevent generating either overly concise responses lacking substantive reasoning or excessively verbose outputs introducing factual errors.

Our full method's superior performance (0.921 Acc. with 0.073 Var.) demonstrates that our carefully balanced reward system creates synergistic effects beyond what individual components achieve in isolation. 
The fact-checking reward becomes most effective when paired with format and length constraints, which provide the structural framework necessary for applying factual evaluation. 
Similarly, semantic similarity becomes more impactful when built upon factually sound reasoning steps. 

\end{appendices}

\end{document}